\documentclass[10pt,journal,compsoc]{IEEEtran}
\ifCLASSOPTIONcompsoc
  \usepackage[nocompress]{cite}
\else
  \usepackage{cite}
\fi
\ifCLASSINFOpdf
\else
\fi
\usepackage{graphicx}
\usepackage{caption}
\usepackage{algorithm}
\usepackage[noend]{algpseudocode}
\usepackage{amsmath}
\usepackage{bbm}

\usepackage{color}
\usepackage{enumerate}
\usepackage{multirow}
\usepackage{float}

\usepackage{changes}
\usepackage{lipsum}
\definechangesauthor[name={Shuo}, color=orange]{sw}

\usepackage{array}

\begin{document}
%
\title{Defeating Misclassification Attacks Against Transfer Learning}
%
%
%
%

\author{Bang~Wu,
        Shuo~Wang,
        Xingliang~Yuan,
        Cong~Wang,
        Carsten~Rudolph
        and~Xiangwen~Yang
\IEEEcompsocitemizethanks{
\IEEEcompsocthanksitem B. Wu, X. Yuan, C. Rudolph and X. Yang  were with the Department of Information Technology, Monash University, Clayton, VIC 3800, Australia. E-mail: \{bang.wu, xingliang.yuan, carsten.rudolph, wayne.yang\}@monash.edu.
\IEEEcompsocthanksitem S. Wang was with CSIRO Data61, Clayton, Melbourne, Australia. E-mail: shuo.wang@data61.csiro.au .
\IEEEcompsocthanksitem C. Wang was with Department of Computer Science, City University of Hong Kong, Hong Kong, China. E-mail: congwang@cityu.edu.hk.
}
\thanks{Manuscript received April 19, 2005; revised August 26, 2015.}
}

%
%

\markboth{Journal of \LaTeX\ Class Files,~Vol.~14, No.~8, August~2015}%
{Shell \MakeLowercase{\textit{et al.}}: Bare Demo of IEEEtran.cls for Computer Society Journals}
%



\IEEEtitleabstractindextext{%
\begin{abstract}
Transfer learning is prevalent as a technique to efficiently generate new models (Student models) based on the knowledge transferred from a pre-trained model (Teacher model). However, Teacher models are often publicly available for sharing and reuse, which inevitably introduces vulnerability to trigger severe attacks against transfer learning systems. In this paper, we take a first step towards mitigating one of the most advanced misclassification attacks in transfer learning. We design a distilled \emph{differentiator} via activation-based network pruning to enervate the attack transferability while retaining accuracy. We adopt an ensemble structure from variant differentiators to improve the defence robustness. To avoid the bloated ensemble size during inference, we propose a two-phase defence, in which inference from the Student model is firstly performed to narrow down the candidate differentiators to be assembled, and later only a small, fixed number of them can be chosen to validate clean or reject adversarial inputs effectively. Our comprehensive evaluations on both large and small image recognition tasks confirm that the Student models with our defence of only 5 differentiators are immune to over 90\% of the adversarial inputs with an accuracy loss of less than 10\%. Our comparison also demonstrates that our design outperforms prior problematic defences.
\end{abstract}

}

\maketitle

\IEEEdisplaynontitleabstractindextext

%
\IEEEpeerreviewmaketitle

\IEEEraisesectionheading{\section{Introduction}\label{sec:introduction}}


\IEEEPARstart{T}{ransfer} learning is widely applied in the development of machine learning systems and has been integrated into commercialised machine learning services, e.g., Google Cloud AI~\cite{GoogleAI}.
%
It facilitates effortless derivation of new models (Student models) through tuning a pre-trained (Teacher model) on a relevant task with less training data and computation cost~\cite{ChenZZ0S020}. 
For example, the InceptionV3 classifier training on ImageNet~\cite{DengJ09} with 1.2 million images requires more than 2 weeks using 8 GPUs~\cite{Szegedy16}. 
By contrast, in transfer learning, a quality face recognition Student model can be developed from a Teacher model with only several hundred training images in several minutes~\cite{Wang18,ChenZZ0S020}.
While expediting the training and curtailing the demands of the data for Student models, transfer learning encounters severe threats~\cite{Wang18,Ji18}, as conventional machine learning systems. 
Well-trained Teacher models become attractive targets, as they are commonly hosted on public platforms and are extensively utilised. 
%
Recent attacks targeting transfer learning~\cite{Wang18} generate adversarial examples to induce misclassification competently.
Based on the boundary conditions revealed from the Teacher models, attackers can imitate and manipulate the input's internal features in the associated Student models, even when these Student models are not accessible to the attackers. 
%
%
Such attacks are powerful, since the transferability of the adversarial manipulation preserves, i.e., no matter whether the structures and parameters of the Student models are modified or not.

To our best knowledge, most of the existing defences against adversarial examples are not suitable in transfer learning. 
For instance, adversarial detection~\cite{Xu0Q18,MaLTL019} loses its ability since the new model structures and weights are modified, and adversarial training~\cite{Tramer17,Kurakin16} based approaches are not optimized for the targeted Student models in transfer learning. 
Wang \emph{et al.} proposed two basic defences for transfer learning, i.e., Randomizing Input via Dropout and Injecting Neuron Distances~\cite{Wang18}, which are either of limited model accuracy or high training difficulty. More importantly, they both fall short of improving the overall robustness of the model, i.e., less effective to non-targeted attacks. 
%
Detailed comparisons can be found in our experiments in Section~\ref{sec:exp}. 

To address these drawbacks, we aim to make a first step towards effectively mitigating the advanced misclassification attacks against transfer learning~\cite{Wang18}.
Our goal is to scalably strengthen the trained customized models (Student models) to detect and reject the adversarial inputs while retaining the model accuracy. 
%

\noindent \textbf{Challenges and Technical Insights:} 
There are several challenges yet to be resolved to achieve the above goal. 
The first is how to reduce the transferability of the attacks.
The targeted Student models are highly vulnerable, since the misclassification attacks feature strong transferability among the Student models that are simply developed from a public Teacher model. 
To break such transferability, we design dedicated classifiers, called differentiators, in a non-trivial manner.
In particular, we carefully apply network pruning to make them highly distilled and vary widely from the Teacher model. 
To expand these dissimilarities, we adapt pruning based on the activation from only two classes, which causes the utmost-possible dissimilarity between the differentiator and the Teacher model. 
%

Each of the differentiators can only mitigate the attacks between two specific classes, and thus employing a single differentiator is not robust to the variance of attacks among other classes.
To overcome this limitation, one can use the ensemble method and group the differentiators with every class pair to cover all possible attacks against arbitrary classes. 
%
Despite being highly effective, we realise that these differentiators developed by only two classes increase the size of the ensemble models for complex classification tasks. 
A practical challenge here is how to reduce long inference latency caused by the bloated ensemble models.

To improve scalability, our observation is that given a certain adversarial example, not all the differentiators are related to the source or target class of it.
Accordingly, we pare our design by selecting differentiators classifying between the preliminary inference class from the Student model and other classes. 
%
The preliminary inference is either the correct source class from a clean example, or the target one from an adversarial example. 
Therefore, the differentiators corresponding to the preliminary inference class are able to validate the clean inputs or reject the attacker's inputs. 
Because a significant portion of our differentiators can successfully reject the adversarial examples, randomly selecting a few of the corresponding ones for the ensemble is sufficient for defence in inference.
%

\noindent \textbf{Contributions:} The main contributions of our work are summarised as follows:
\begin{itemize}
\item To our best knowledge, we are the first to propose effective yet comprehensive defences for transfer learning. 
Our design mitigates both the targeted and non-targeted misclassification attacks~\cite{Wang18}. The former attacks generate adversarial inputs which can be identified to a target class, while the latter ones can misclassify the inputs to any other classes. 
\item We carefully utilize network pruning to build differentiators, where each differentiator is designed to infer two specific classes dedicatedly and to be immune to adversarial inputs. 
To deduct the transferability of the attacks, we apply activation pruning when developing the distilled models.
To preserve the accuracy, we choose flexible pruning ratios for different layers to reduce the accuracy loss caused by pruning, and iteratively retrain the pruned models.
For efficiency, we further adopt independent pruning for each layer to curtail the time cost of the design implementation.
\item We incorporate an ensemble structure into our defence to improve the robustness of the overall models against the attacks among all classes.
%
We instantiate our ensemble structure using a general developing Student model and a group of our two-class differentiators. 
%
In the first phase, the preliminary inference result is utilised from the Student model to narrow down the possible source or target classes of the inputs.
In the second phase, only a small, fixed number of the differentiators that correspond to these classes validate clean or reject adversarial inputs.
As a result, our design satisfies the defence rate, model accuracy, and scalability.
\item We implement our defence for two transfer learning applications, i.e., Face Recognition (83 classes) and Traffic Sign Recognition (43 classes). 
We evaluate our design on the defence rate, model accuracy, efficiency of the model development and inference, and even effectiveness against general attacks like FGSM~\cite{Goodfellow14} and DeepFool~\cite{Moosavi16}. 
The results for both tasks confirm that our ensemble models can reject over 91\% of the adversarial examples only with 5 differentiators in 2 seconds. 
It achieves more than 90\% defence rates with different attack configurations on attack layers and perturbation budgets. 
Meanwhile, our design preserves the accuracy above 90\%  after decent pruning and retraining. 
Last but not least, we conduct comprehensive comparisons with prior arts to demonstrate that our defence achieves higher defence rates, particularly for non-targeted attacks. 
Our defence rates are $\sim$90\% for both tasks, while prior arts proposed in~\cite{Wang18} only reach $\sim$20\% for Face and $\sim$40\% for Traffic Sign Recognition. 
\end{itemize}
\noindent \textbf{Organisation:}
The rest of our paper is organised as follows. 
Section~\ref{sec:background} provides the background knowledge of misclassification attacks in transfer learning and network pruning used in our design.
%
Section~\ref{sec:defence} presents our defence design in detail.
Section~\ref{sec:exp} shows the experimental results of our defence and the comparisons among our design to others. 
Section~\ref{sec:discussion} discusses the effectiveness of our design in broader scenarios.
Section~\ref{sec:relatedwork} introduces related work on attacks and defences of machine learning and transfer learning systems.
Section~\ref{sec:conclusion} gives a conclusion and future directions.

\section{Background}
\label{sec:background}
\subsection{Misclassification Attacks against Transfer Learning}
%

\noindent \textbf{Transfer Learning:}
Transfer learning is proposed to learn knowledge from a completed model and prediction task, and improve the training of new models for different tasks.
The knowledge can be either domain information, which consists of the feature space and the marginal probability distribution of the training data, or the learning task, which consists of a label space of the training data and the pre-trained model fitting the objective predictive function of this task~\cite{ChenZZ0S020}.
%
%
Based on existing knowledge, transfer learning speeds up the development of the new models even when their domains or learning tasks are different.
%
%
%
%
%
\begin{figure}[!t]
\centering
\includegraphics[width=0.45\textwidth]{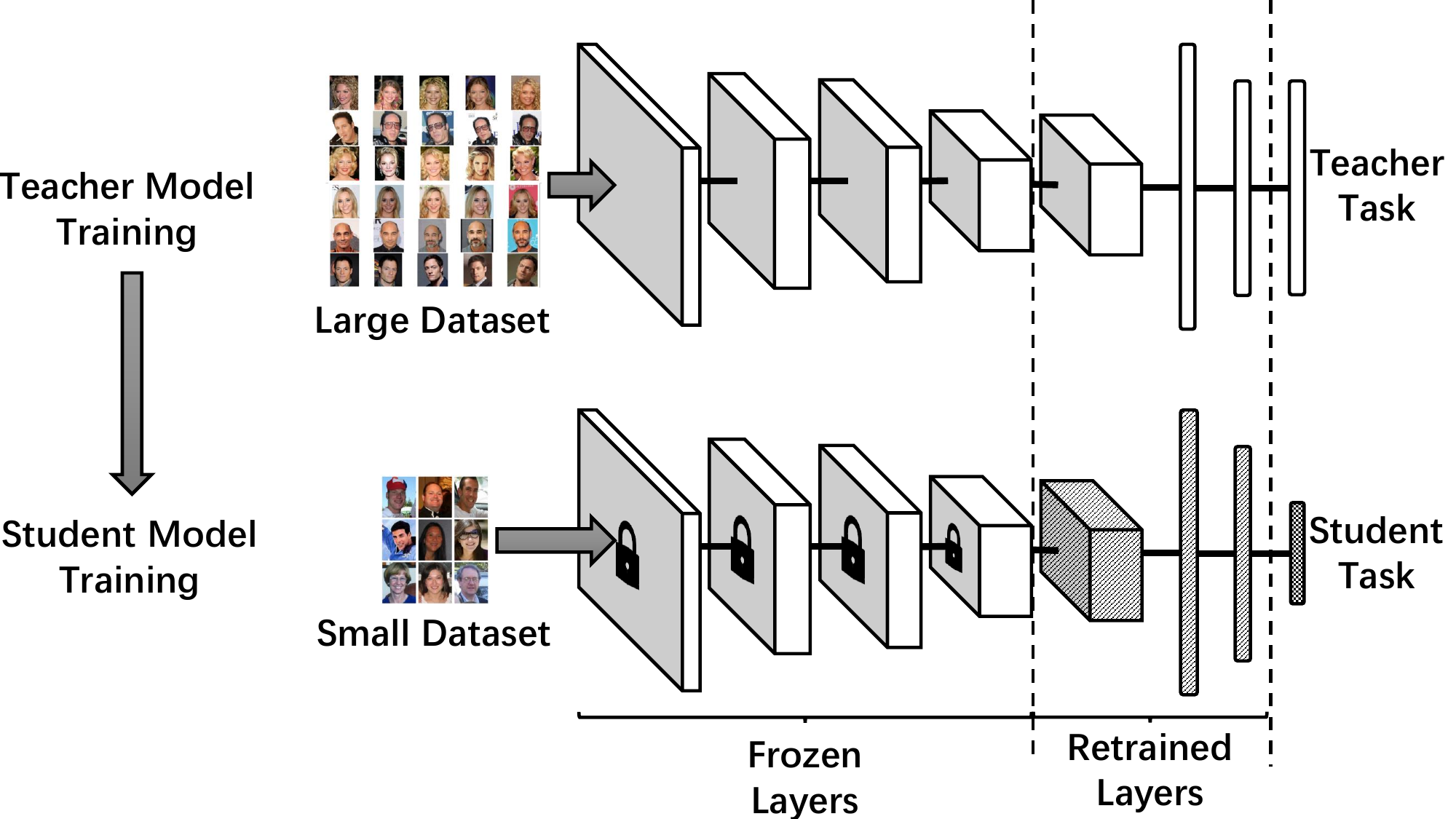}\\
        \caption{Transfer Learning}
        \label{fig:TransferLearning1}
\vspace{-10pt}
\end{figure}

A simple way of transfer learning is developing a new model based on both weights and architectures of the layers from a well-trained model.
If the new model has a similar domain or learning task as the pre-trained model, it can be directly built by fine-tuning the parameters to fit its task.
For a transfer learning process, the Student model first copies both the architecture and weights from the Teacher model. After that, the last classification layer of the Student model is tailored to fit the new classification task. Then, the Student model is tuned based on the similarity of the two tasks. One common methodology of tuning is to freeze several layers and retrain the rest of them as shown in Figure~\ref{fig:TransferLearning1}. 

As we mentioned before, reusing the Teacher model introduces the vulnerability in the Student models. 
Recent studies have exploited it and propose adversarial attacks specifically targeting transfer learning systems. 
To our best knowledge, the most effective and easily deployable ones are the misclassification attacks introduced by Wang \textit{et al.}~\cite{Wang18}.
%

\noindent \textbf{Attacks Assumption:}
To be consistent, we follow the same assumption as Wang \textit{et al.}~\cite{Wang18}. 
The attacks assume white-box access to Teacher models and black-box to Student models.
%

\noindent \textit{White-box Teacher Model:}
We first assume that the attackers have full access to the Teacher model. This is realistic because most of the well-trained models are publicly available. The attackers can pretend to be one of the students so that they can know both the weights and the architecture of the Teacher models. We also assume that the attackers can find the corresponding Teacher model when they are targeting a Student model. ~\cite{ Wang18} introduces this Teacher model fingerprinting method. 

\noindent \textit{Black-box Student Model:}
The attackers are assumed to have no access to the Student models. Namely, the Student models are considered as black boxes. We assume that neither the model parameters (including the weights and the architecture) nor the training datasets for Student models are accessible to the attackers. In real situations, this information may include sensitive and private data that is normally considered proprietary to Student model owners. 
Besides, we assume that the attackers can only use limited queries to the implemented Student models, which makes them hard to reproduce the shadow models.

To make coherent assumptions as~\cite{Wang18}, we do not consider the case where the Student models are reproduced or leaked. 
If the attackers can directly gain enough information from the Student models rather than from the Teacher models, the implementation of the attacks will not be impacted by the transfer learning method. 
Namely, this case will become a generic attack and defence problem in machine learning. 
Nevertheless, our design further considers a stronger attacker according to a guiding principle introduced in~\cite{Carlini19}, which suggests that the adversaries might obtain some knowledge of the defence algorithm. 
More details can be found in Section~\ref{sec:discussion}.

\begin{figure}[!t]
\centering
\begin{minipage}[htp]{1\linewidth}
        \centering
        \includegraphics[width=1\textwidth]{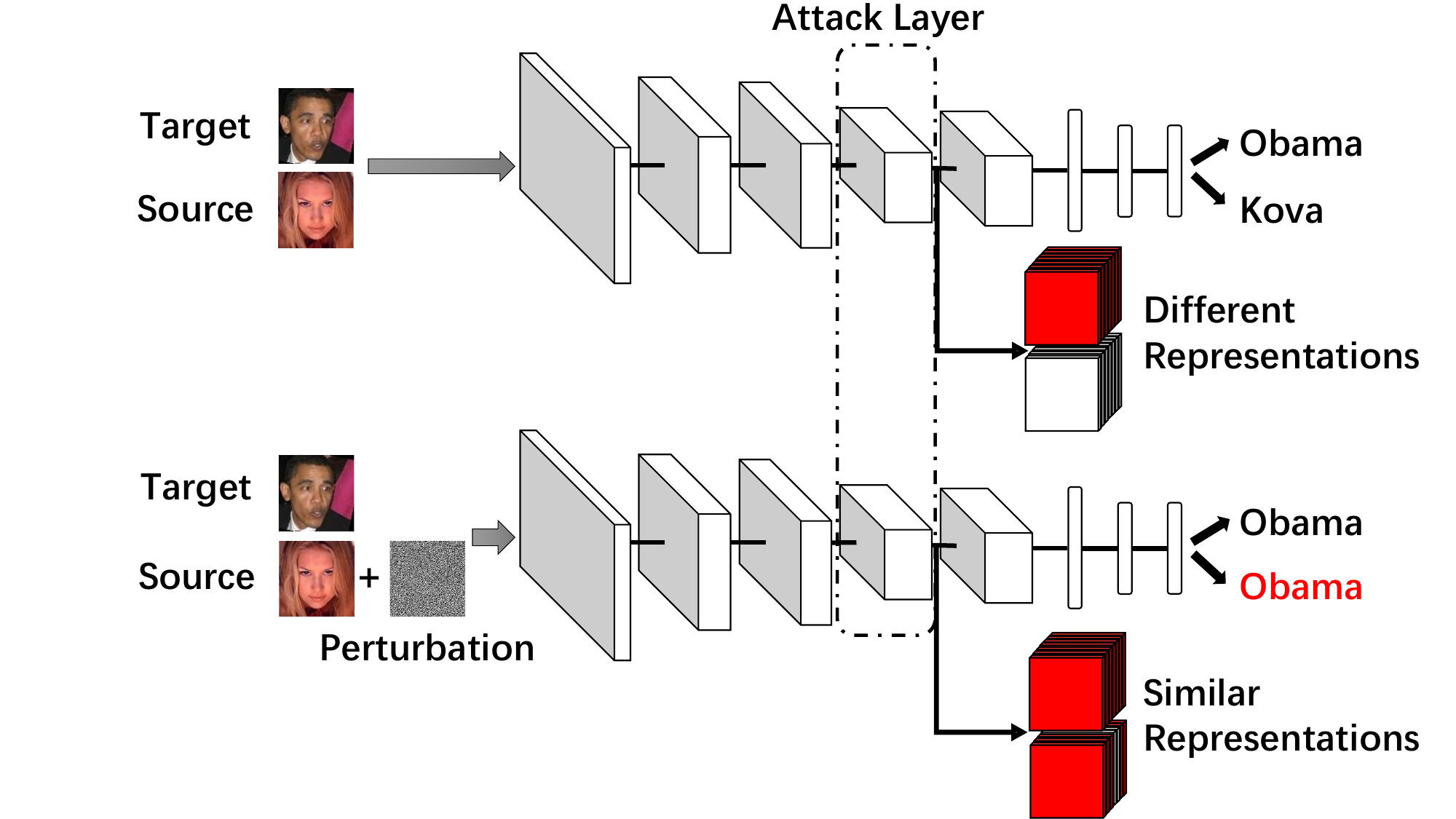}
        \caption{Misclassification Attacks in Transfer Learning~\cite{Wang18}}
        \label{fig:TransferLearningAttack1}
\end{minipage}
\vspace{-10pt}
\end{figure}

\noindent \textbf{Misclassification Attacks Methodology:}
Figure~\ref{fig:TransferLearningAttack1} depicts the idea of how the misclassification attacks~\cite{Wang18} conduct in transfer learning.
%
Given a transfer learning system, the attackers expect a carefully chosen layer where the layers before it may be frozen or just be slightly tuned during the development of the Student models. 
Thus, the attacks can make the internal features of a certain layer output in Teacher models for two different input images be very similar. 
%
Such similarity will be maintained to the final prediction.
As a result, the misclassification takes place since two inputs with different labels have a similar prediction.
%

%
For realisation, the misclassification attacks in transfer learning are minimising the distance for the internal outputs within perturbation budgets.
%
%
Specifically, the misclassification attacks can be translated to an optimised problem as equation~\ref{equ:TargetedAttack1}.
\begin{equation}\label{equ:TargetedAttack1}
\begin{split}
    &\min D(H_K(x'_s),H_K(x_t))\\
    &s.t. \quad d(x'_s,x_s)<P
\end{split}
\end{equation}
The distance (measured by $D(.)$) between both internal feature $H(.)$ at attack layer $K$, for the adversarial input $x'_s$ and the target inputs $x_t$ is minimised.
The perturbations adding to source inputs $x_s$ is limited by the budget $P$ (measured by distance function $d(.)$).
%

The misclassification attacks can be both targeted and non-targeted. 
A targeted attack is applied by choosing a mimicked input with a targeted label class and construct the adversarial examples by solving the optimisation function in equation~\ref{equ:TargetedAttack1}. 
A non-targeted attack is applied by evaluating multiple adversarial images for different targeted attacks and choosing the one with the smallest internal feature dissimilarity.
%
%
%
%
According to~\cite{Wang18}, a subset including five supposed targeted attack images is sufficient to find adversarial images with high attack success rates.

%
\subsection{Network Pruning} 
\begin{figure}[!t]
\centering
\includegraphics[width=0.4\textwidth]{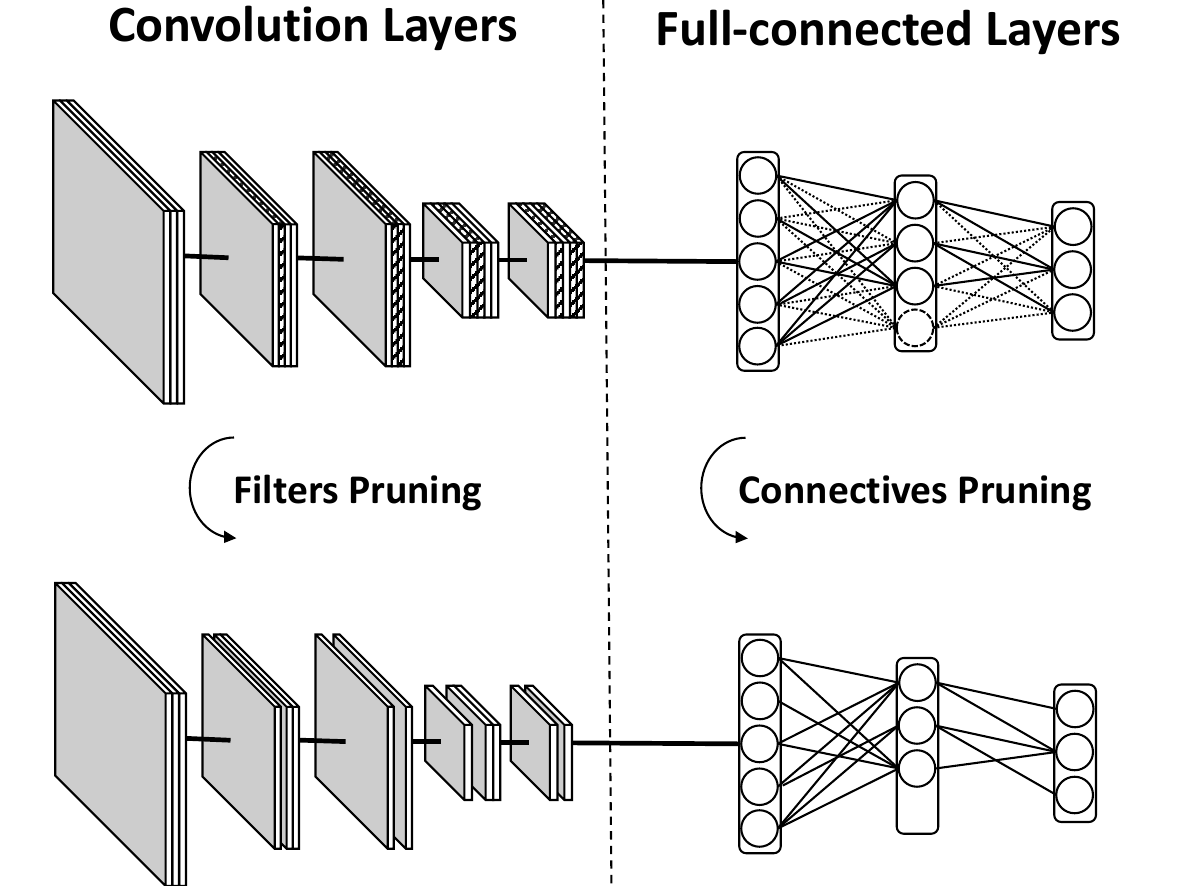}\\
        \caption{Network Pruning}
        \label{fig:Pruning1}
\vspace{-10pt}
\end{figure}

%
Network pruning aims to remove the unimportant connections of a network which makes a dense neural network become a sparser one.
By carefully choosing the pruned connections and further tuning the networks, the accuracy loss of the pruned networks can be acceptable.
%
According to prior work~\cite{Han15,Wang19,LiuSZHD19,MolchanovAV17}, most of the model structures have redundant neurons and connectives.
These connectives are less active during the classification tasks.
Pruning these unnecessary components can improve the efficiency of both inferring and storing for the machine learning models. 
Figure~\ref{fig:Pruning1} overviews the pruning over a three-layer neural network.

In particular, there are two approaches for pruning.
\textbf{(1) Weight-based pruning.} It is a simple and direct way.
A threshold weight value can be chosen and all connectives with fewer weight values than the threshold can be removed~\cite{Han15}.
\textbf{(2) Activation-based pruning.} It considers how the weights are activated by the expected inputs~\cite{Polyak15}.
It removes the connectives with fewer activation compared to the threshold.
Besides, for large neural networks with convolution layers, the convolution matrix (also called kernel or filter) rather than each connective can be pruned.
It is shown that removing entire filters can produce a network with a more regular structure compared to remove ~\cite{Han15,Polyak15}.

After pruning, a further process should be applied to reduce the loss of accuracy.
A simple way is to retrain the pruned network.
In prior work~\cite{Han15,Polyak15}, in order to achieve a high pruning rate with less accuracy loss, the network is pruned and retrained iteratively by gradually increasing the pruning rates.
\section{Defence Design}
\label{sec:defence}
\subsection{Defence Goal and Evaluation}
\noindent \textbf{Goal:} Our goal is to develop a defence approach to address the attacks described above.
%
Considering the real application scenarios for the defenders, several assumptions are made. 
Since robust machine learning systems against the adversarial inputs are desired for users, our defence's development can be fully supported by the customers owning the Student models.
%
Naturally, the defenders are assumed to have access to both weights and structures of the Student models as well as their training data.
They are also assumed to be able to modify the Student models and make them more robust to the adversarial images.


\noindent 
\noindent \textbf{Evaluation:}
In particular, the defence will be evaluated in terms of both efficiency and effectiveness as follows:
%

\noindent \textit{Classification Accuracy:} The models with our defence should be able to classify most of the clean images successfully. 
The adversarial examples are still less possible to happen, it is necessary to maintain the prediction accuracy on the clean inputs. 

\noindent 
\textit{Defence Success Rate:} 
Our defence is expected to detect most of the adversarial inputs for both the targeted and non-targeted and reject them.
%
%
%

\noindent 
\noindent \textit{Time Consuming:} As one of the motivations of transfer learning is to save cost for large scale learning tasks, our defence should introduce a comparable inference time cost to the ordinary inference or prior other defences. 

\noindent 
\noindent \textit{Model Size:} Our defence is also expected to consume acceptable storage resources. 
Our defence should be scalable for large models, because transfer learning will be widely adopted when the Teacher models become large and complex. 

\begin{figure}[!t]
\centering
\begin{minipage}[!t]{1\linewidth}
        \centering
        \includegraphics[width=1\textwidth]{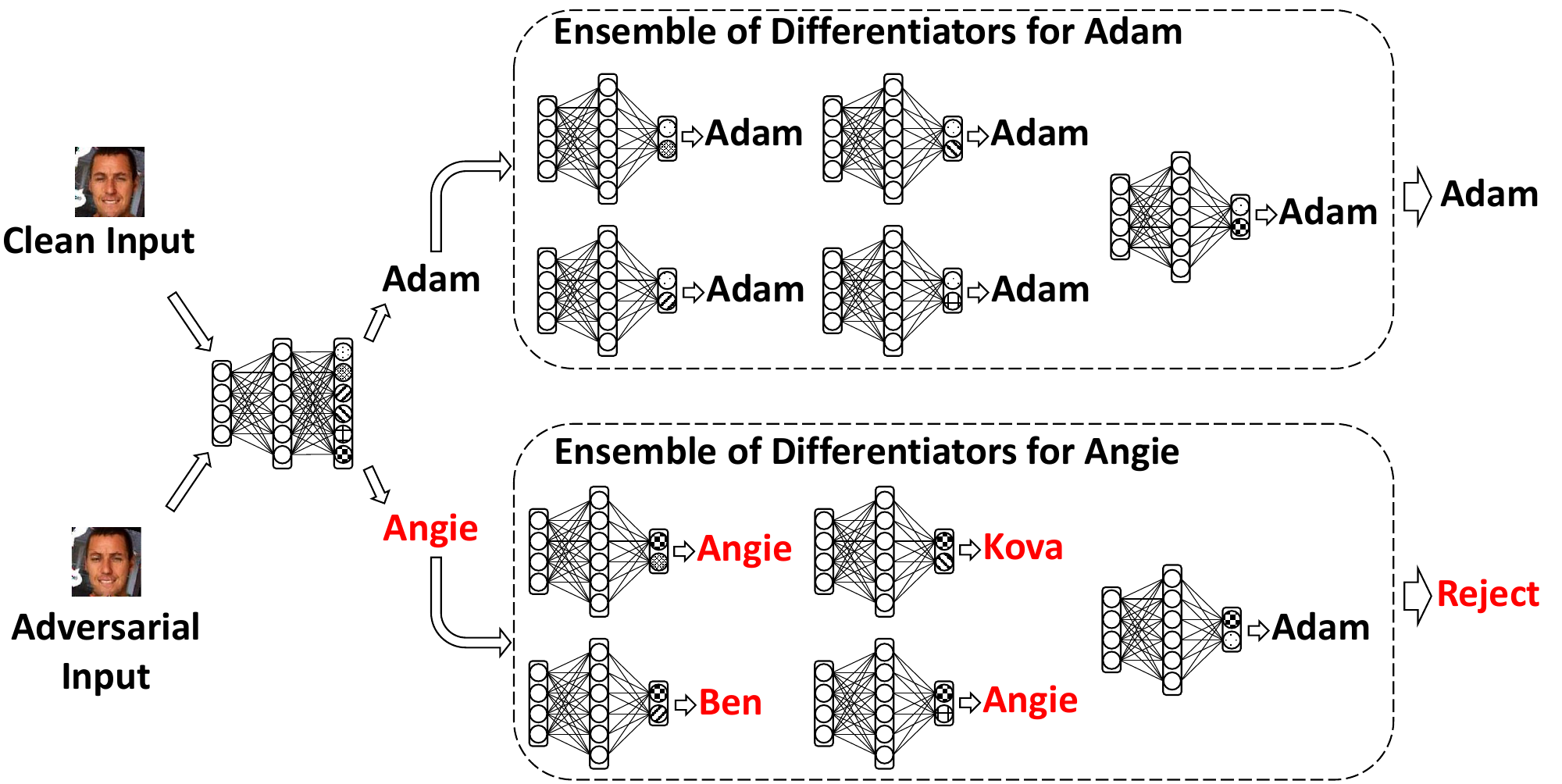}\\
        \caption{Inference Strategy}
        \label{fig:EnsembleStructure11}
\end{minipage}
\vspace{-10pt}
\end{figure}
\subsection{Defence Intuition}

%
%
%
We first present the design intuition of our defence against misclassification attacks.
As introduced in Section~\ref{sec:background}, the effectiveness of the misclassification attacks relies on manufacturing a similar internal representation at an attack layer.
To address these attacks, a direct and effective solution is modifying the models to make them different from the original ones.
Among others, network pruning has been shown effective in modifying the model structure and restrain the transferability of the attacks~\cite{Liu18,Wang18}. 
As introduced in Section~\ref{sec:background}, pruning removes lots of connections of the models, which makes them far from the original dense ones.
The difference between the pruned models and targeted models can reduce the effectiveness of the attacks.
Besides, the sparse networks are also more efficient for development (fine-tuning) and application (inference of the models) later, which conforms to the advantages of using transfer learning.

To further reduce the attack success rate while extending the defence to non-targeted attacks, we adapt ensemble models, which are widely used in a machine learning system.
The intuition is that, with limited perturbation, it is hard for the adversarial inputs to maintain the transferability to the variance of models.
The adversarial images can only fool parts of these models and lose their transferability to the rest of them.
The overall ensemble models with different classifiers achieve stronger robustness against these attacks.

While extending the defence to non-targeted attacks, a naive approach that combines differentiators for all possible attack pairs results in a large ensemble model, which contains $(K-1)K/2$ differentiators for a $K$ label classification problem).
To improve the scalability of our design and reduce the inference latency caused by the large size of the ensemble models, we design a two-phase inference.
The inputs are first predicted by the Student model trained by general transfer learning without pruning.
The inference result should be either the correct source class from a clean example or the target one from an adversarial example.
Therefore, differentiators classifying between the source or target class and other classes are chosen to validate or reject the inputs.
The distinctiveness among these differentiators and the Student model is sufficient to reduce the attack effectiveness due to the activation pruning based on only the activation of their corresponding classes. 
As a result, a small group of randomly selected differentiators can still be different, making it hard for the adversarial inputs to fool all of them.
The ensemble models consisted of them can be robust to the adversarial examples. 

\noindent 
\textbf{Remark:}
Our approach combines activation pruning with ensemble models to mitigate the transfer learning attacks. 
Note that simple pruning alone will not be sufficient to mitigate the advanced transfer learning attacks. 
Our observations suggest that the neurons activated by the adversarial examples overlap with those activated by the clean samples, therefore pruning the general student model is not sufficient to mitigate the advanced transfer learning attacks while maintaining model accuracy.  
Our design of the differentiators enables them to be more efficiently pruned compared to the general classifier. 
Since our differentiators are used only to distinguish between two labels, the number of neurons activated by their corresponding inputs can be reduced considerably, which facilitates a thorough pruning of the model. 
Accordingly, these highly distilled models can effectively defend against adversarial examples while remaining accurate for clean inputs. 
Further, we propose ensemble models that combine the differentiators in order to mitigate targeted attacks with different source and target labels as well as non-targeted attacks.  
As a result, it is necessary to combine such two techniques in order to mitigate the attacks we wish to prevent, whereas prior research using only one of them cannot be effective. 
\subsection{Defence Implementation}
This section introduces the details of our proposed defence implementation. 
Specifically, we will describe how to construct the ensemble models and then develop the differentiators. 

\subsubsection{Ensemble Construction}
\label{subsubsec:ensemble}
\begin{algorithm}[!t]
\caption{Inference Strategy}\label{alg_defence1}
\begin{flushleft}
\hspace*{\algorithmicindent}\textbf{Input:} \\
\hspace*{\algorithmicindent}$x$, input image; \\
\hspace*{\algorithmicindent}$S_o$, general Student model; \\
\hspace*{\algorithmicindent}$S_{1,2}$, $S_{1,3}$, ... $S_{K-1,K}$,  differentiators classifying between
\hspace*{\algorithmicindent} every two classes in a K classes classification problem,
\hspace*{\algorithmicindent} their outputs are labels from 1 to K;\\
\hspace*{\algorithmicindent}$k$, size of the ensemble models, where $k<K$. \\
\hspace*{\algorithmicindent}\textbf{Output:} \\
\hspace*{\algorithmicindent}$y$, final prediction.
\end{flushleft}
\begin{algorithmic}[1]
\Function{inference}{$x, S_o, S_{1,2}, S_{1,3}, \dots S_{K-1,K}$} 
    \State $//$ Phase 1: preliminary inference
    \State $pre\_result \gets S_o(x)$
    \State $//$ Phase 2: validate clean or reject adversarial inputs
    \State Obtain $\{S_{i,j}\}$, where $pre\_result$ is $i$ or $j$
    \State Randomly select $\{S\}_{k}$ from $\{S_{i,j}\}$
    \For {each $S \in \{S\}_k$}
        \State $valid\_result \gets S(x)$ 
        \If{$valid\_result \neq pre\_result$}
            \State $//$ reject adversarial input defined as class 0
            \State \Return{0} 
        \EndIf
    \EndFor
    \State $//$ clean input passes the validation
    \State $y \gets pre\_result$
    \State \Return{$y$}
\EndFunction
\end{algorithmic}
\end{algorithm}

The inference consists of two steps: a preliminary inference by the original Student model and a validation inference by a group of differentiators.
For a classification problem with $K$ classes, the inputs are first predicted by the general Student model.
For any inference results as class $c_i$, there are $K-1$ possible label pairs $(c_i,c_j)$, where each of them corresponds to a differentiator trained by their corresponding subsets.
%
%
%
%
%
%
%
%
They are expected to be fully pruned and robust to the adversarial inputs according to the design goals of the differentiators.
%

Figure~\ref{fig:EnsembleStructure11} illustrates an example of the above progress.
For any clean inputs, the Student model is likely to provide a correct preliminary prediction.
All the differentiators then can validate this result as they are trained by a subset of training data consisted of the correct label dataset.
For an adversarial input, the Student model might be easily fooled and provide a wrong prediction.
However, it is difficult for the adversarial input to fool all differentiators.
If one of the differentiators comes to a different prediction, the input is considered as an adversarial example and rejected.

To reduce the latency of our design, our observation is that a small number of randomly selected corresponding differentiators are sufficient to reject almost all the adversarial images. 
%
%
Besides, inferring among more differentiators may reduce the accuracy of the whole ensemble models due to False Negative.
As a result, our implementation aims to minimise the number of differentiators in regard to the defence specification of the applications.
The detailed strategy is described in Algorithm~\ref{alg_defence1}. 
Later in our experiments, 5 differentiators achieve balance on defence rate and accuracy for both small and large tasks.
\subsubsection{Distilled Differentiator}
In our design, network pruning is applied to develop the distilled differentiators.
To reduce the accuracy loss of pruning, they are pruned based on the activation of their whole training data, which is shown in Algorithm~\ref{defence2}.
\begin{algorithm}[!t]
\caption{Differentiator Generator}\label{defence2}
\begin{flushleft}
\hspace*{\algorithmicindent}\textbf{Input:} \\
\hspace*{\algorithmicindent}$D_i,D_j$, two training datasets for the differentiator; \\
\hspace*{\algorithmicindent}$S_{teacher}$, the Teacher model of the transfer learning task. \\
\hspace*{\algorithmicindent}$act$, the activation of the two class for Teacher model. \\
\hspace*{\algorithmicindent}\textbf{Output:} \\
\hspace*{\algorithmicindent}$S_d$, a distilled differentiator.
\end{flushleft}
\begin{algorithmic}[1]
\Function{DifferentiatorTraining}{$D_i, D_j, S_{teacher}$}
\For{each $layer$ in $S_{teacher}$}
    \If{$layer$ is convolution layer}
        \State Filter Pruning in $layer$ based on $act$
    \EndIf
    \If{$layer$ is full-connected layer}
        \State Connective Pruning in $layer$ based on $act$
    \EndIf
\EndFor
\State $S_{student} \gets TransferLearning(D_i, D_j, S_{teacher})$
\For{$i=1$ to $IterationTimes$}
    \For{each $UnfrozenLayer$ in $S_{student}$}
        \State Pruning in $UnfrozenLayer$ based on $act$
    \EndFor
    \State $S_d \gets FineTuning(S_{student})$
\EndFor
\State \Return{$S_d$}
\EndFunction
\end{algorithmic}
\end{algorithm}

As we introduced above, each differentiator is trained to classify between two classes by its corresponding dataset.
%
To save the training cost, our design applies the general transfer learning method to build the differentiators.
As introduced in Section~\ref{sec:background}, some of the layers copied from the Teacher model are frozen, and the rest of them will be retrained by the dataset for the differentiator's tasks. 
%
After that, these differentiators are distilled via pruning each layer. 
The details are presented as follows.

\noindent \textbf{Activation Pruning:}
We use activation pruning which has been demonstrated comprehensive by considering the effect of both inputs and models parameters.
In our design, the differentiators are expected to be highly distilled and variant to each other.
The activation pruning based on different training data increases the disparity among these differentiators, which can improve the robustness of ensemble models.
Meanwhile, activation pruning is also shown to have less accuracy loss compared to weight pruning.
Therefore, the distilled models via activation pruning are more adaptive to the Student model tasks.
Note that our differentiators only focus on the corresponding two classes.
Therefore, they can still achieve high accuracy, even pruned and tuned by a small dataset, which is desired in transfer learning scenarios.

\noindent \textbf{Pruning via Ratio:}
Our design prunes the models via ratio rather than threshold values.
Based on our observation, the values of the activation can be entirely different for each differentiator while the pruning ratios can be limited to a small range, which expedites our defence development.

\noindent \textbf{Different Ratio for Different Layers:}
In our defence, different pruning rates are chosen for different layers in each differentiator.
According to previous work~\cite{Li16}, each layer has a different sensitivity corresponding to the final model accuracy.
Therefore, pruning ratios for different layers in our design correspond to their pruning sensitivity~\cite{Han15}.
They are chosen following general pruning strategies~\cite{Han15,Polyak15} to maintain the overall accuracy while pruning the redundant components as much as possible.
Since each differentiator is pruned based on activation, using similar pruning ratios still results in contrasting models.
Based on our experiments, more than half of the differentiators can share the same pruning ratios while achieving a high defence rate. 
As a result, the efforts of tuning pruning rates are somehow limited.
%
%

\noindent \textbf{Filters and Connectives:}
To distil the differentiators better, we apply diverse strategies for different types of layers.
For the full-connected layers, we prune every single connectivity evaluating their activation.
For the convolution layers, we prune the filters consisting of correlative connectives.
While the work~\cite{Li16} shows that pruning the filters in the convolution layers makes the networks more efficient, we find it also improves the defence.

\noindent \textbf{Independent Pruning:}
We prune each layer separately, where the pruning for each layer is not affected by the others.
By pruning each layer independently, every layer can be pruned in parallel which makes our pruning more efficient.  
The activation of each class for activation pruning are calculated once and reused when developing other differentiators.
%


\noindent \textbf{Iteration Pruning and Retraining:}
To preserve accuracy, we propose to retrain and prune the last several layers of the models iteratively.
As directly pruning the networks will harm the accuracy of the models, the classifiers can regain accuracy by performing iteration pruning and retraining the whole models~\cite{Han15}. 
To limit the computation cost, the iterative method is only applied to the unfrozen layers.


\section{Experimental Results}
\label{sec:exp}
In this section, we report the experimental results of our defence. 
We demonstrate that our design highly improves the robustness of the models against the misclassification attacks in transfer learning.
We also show that our defence is scalable for various classification tasks and adversarial examples with different configurations.
Specifically, we evaluate our defence in accuracy and defence success rates, and compare it with prior arts over two typical tasks. 
We also evaluate a popular attack in conventional machine learning systems, i.e. FGSM~\cite{Goodfellow14} to show that our design is applicable to general attacks of adversarial samples. 
%
\subsection{Experimental Setup}
This section introduces the setups of our experiments. Specifically, we will propose the application tasks, the transfer learning scenarios and then the attack setup.
\subsubsection{Teacher and Student Models Selection}

To evaluate our defence, we apply the misclassification attacks to two popular transfer learning tasks: (1) Face Recognition (recognising 83 public persons in a dataset called PubFig~\cite{Pinto11}), and (2) Traffic Sign Recognition (recognising 43 traffic signs in a dataset called GTSRB~\cite{Stallkamp11}).

%

\noindent \textit{Face Recognition:} The task is to classify human faces.
It is a common task used to evaluate both attacks and defences.
The Teacher model is a popular and public pre-trained model, called VGG-Face~\cite{Parkhi15} which is well trained by 2.6M faces with an accuracy of over 90\%. 
The Student model will be trained to classify 83 persons chosen from the PubFig dataset~\cite{kumar09} with only 7470 images.
%


\noindent \textit{Traffic Sign Recognition:}
This task is to classify different traffic signs for an auto-driving system.
The Teacher model is a normal VGG16 model~\cite{Simonyan14} trained via the ImageNet dataset with 14 million images.
%
The top-5 test accuracy of this pre-trained model is about 90.1\%.
%
The training data for our Student model comes from the GTSRB dataset~\cite{Stallkamp12} which includes 39209 images of 43 traffic signs.
As mentioned in Section~\ref{sec:background}, transfer learning is commonly used with benefits when developing a Student model with relatively small-scale training data. 
To simulate this scenario, both Teacher models used in our experiments are popular pre-trained models derived from a large amount of data, while both Student models are using the much smaller-scale training datasets comparing the Teacher models.
%
%
%
%


\subsubsection{Transfer Method Selection}
In particular, there are three approaches for the realisation of transfer learning based on the extent of the tuning~\cite{Wang18}. 
One is called Deep-layer Feature Extractor, where only the last layer is changed and trained by the Student model training datasets. 
The second one is a Mid-layer Feature Extractor that unfreezes and retrains some of the layers. 
The third one named Full Model Fine-tuning unfreezes and tunes all of the layers.

According to previous work~\cite{Wang18}, the misclassification attacks are less effective for the Student models developed using Full Model Fine-tuning.
Therefore, only the applications for the former two approaches are evaluated in our experiments. 
For the first task, both the Teacher model and the Student model focus on face recognition. 
Therefore, the transfer learning system can be applied in a direct and simple way.
Based on the observation in~\cite{Wang18}, a Face Recognition model developed as a Deep-layer Feature Extractor achieves higher accuracy than other transfer processes.
As a result, the Deep-layer Feature Extractor is built for Face Recognition in our experiments.
For the second task, the Teacher model is classifying general objects, while the Student model focuses on traffic signs. 
Therefore, we again follow the same configuration as~\cite{Wang18} and use Mid-layer Feature Extractor in this task.
The cut-off layer is chosen to be layer 10 out of 16 for the VGG16 Model when developing the Student model to achieve high accuracy.
%
%

\subsubsection{Attack Setup}
In our experiments, we generate the adversarial images following the same steps as~\cite{Wang18} as discussed above which are sufficient to evaluate our defence. 

\noindent \textbf{Attack Pairs:}
Both the source and target images are randomly chosen from the test dataset, and they are not used for training the Student model. 
This treatment matches the assumption that the Student model is black-box, and the attacker cannot obtain the training data.
%
For the targeted attacks, we randomly choose 1K source and target image pairs to generate adversarial images.
%
For the non-targeted attacks, we also generate 1K adversarial images by randomly choosing source images and 5 target images with different classes for each of them.
After that, we evaluate the distance between the internal feature vectors of the adversarial and target images. 
The source and target pair with the smallest internal representation distance will be chosen to generate the final adversarial images of the non-targeted attacks.

\noindent \textbf{Attack Configuration:}
The adversarial images are generated to target different attack layers.
The optimal attack layer with the highest attack success rate will be considered as the final attack layer~\cite{Wang18}. 
%
The perturbation budget of the adversarial images is $0.003$ in the DSSIM metric~\cite{Wang18} for the Face Recognition task and $0.01$ for the Traffic Sign Recognition task, which is considered as less detectable thresholds~\cite{Wang18}. 
DSSIM is a distance matrix evaluating the structural similarity of images~\cite{Wang18}.
It evaluates the difference between two input images similar to the human's criterion which is suitable for image recognition.
%
We use Adadelta~\cite{Zeiler12} as the optimiser of the adversarial sample generator. 
The optimised problem of the adversarial images generation uses $2000$ for the iteration times and $1$ for the learning rate.


\subsection{Evaluation}

We evaluate our design by comparing the robustness of the systems with and without our design. We also compare our design with other defences to demonstrate the advantages.
In particular, we appraise both the effectiveness and efficiency of our design. 
Regarding the effectiveness, we evaluate metrics, including our defence rate against the attacks targeting at different layers.
We also demonstrate the defence rate in the perturbation budget of the attacks and show that our design is effective for variance attack configurations.
Besides, we indicate the accuracy preservation of our defence by showing the relationship between iteration number and model accuracy for the clean inputs.
In addition, we apply other attacks to our design and show that our design methodology can readily be deployed to address general attacks in machine learning systems.
Regarding efficiency, we evaluate the memory and time cost of our design.
Finally, we compare benign input accuracy, defence performance, and efficiency of our design with prior defences in~\cite{Wang18}.




\begin{figure*}[!t]
\centering
\begin{minipage}[htp]{0.31\linewidth}
        \centering
        \includegraphics[width=0.95\textwidth]{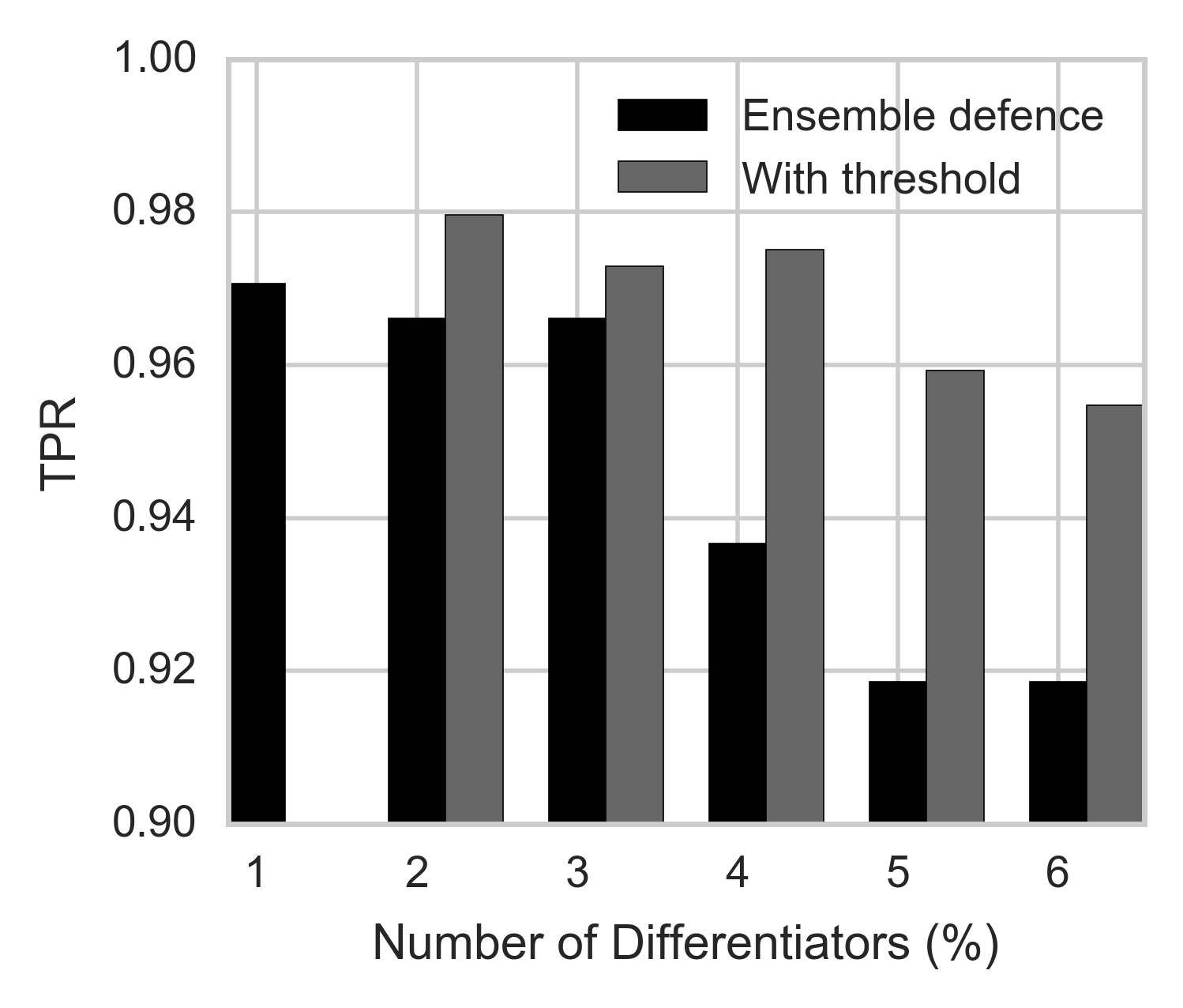}
        
        (a) Face Recognition
        
\end{minipage}
\begin{minipage}[htp]{0.54\linewidth}
        \centering
        \includegraphics[width=0.95\textwidth]{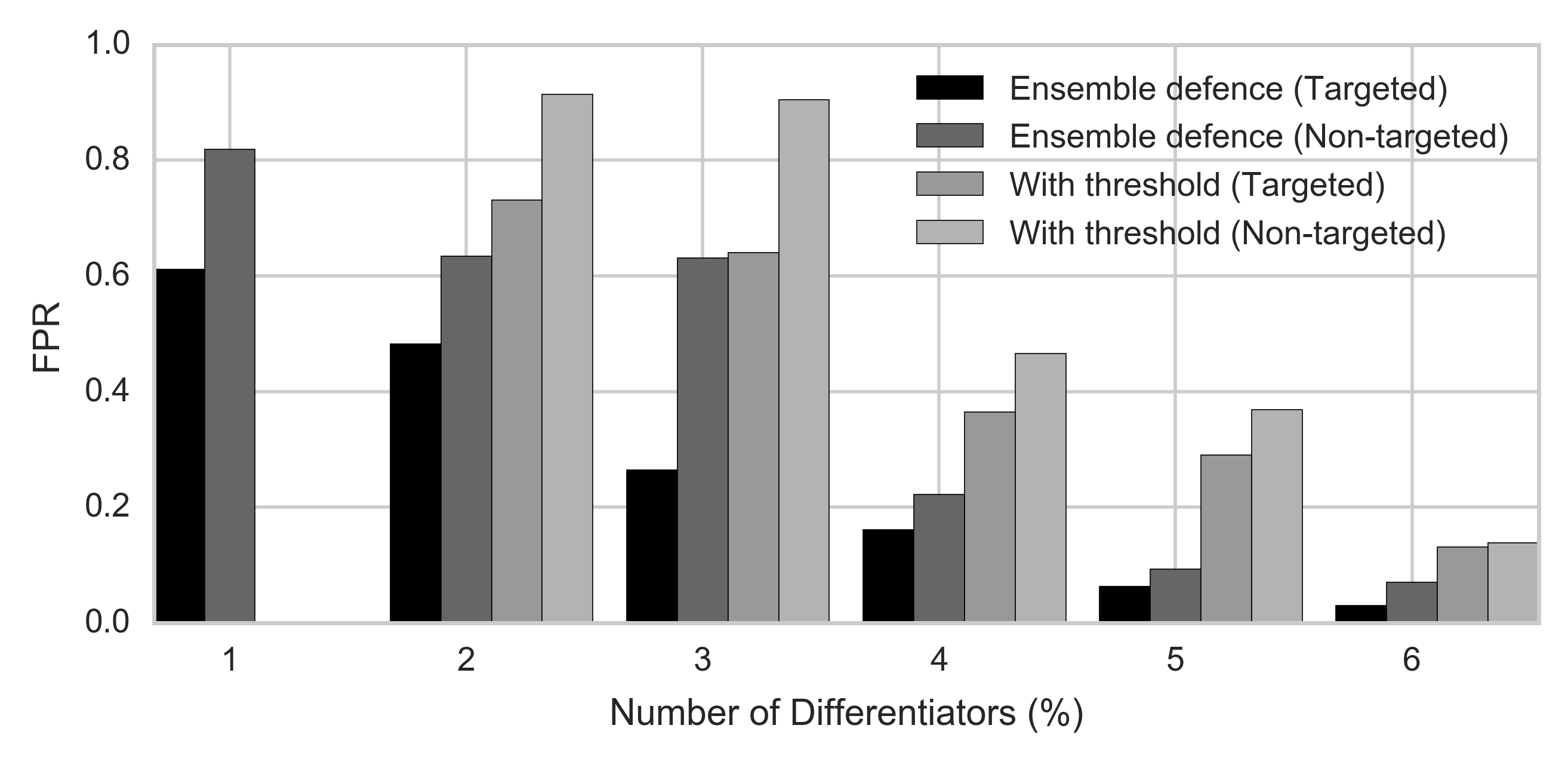}
        
        (b) Traffic Sign Recognition
        
\end{minipage}
\\
\begin{minipage}[htp]{0.31\linewidth}
        \centering
        \includegraphics[width=0.95\textwidth]{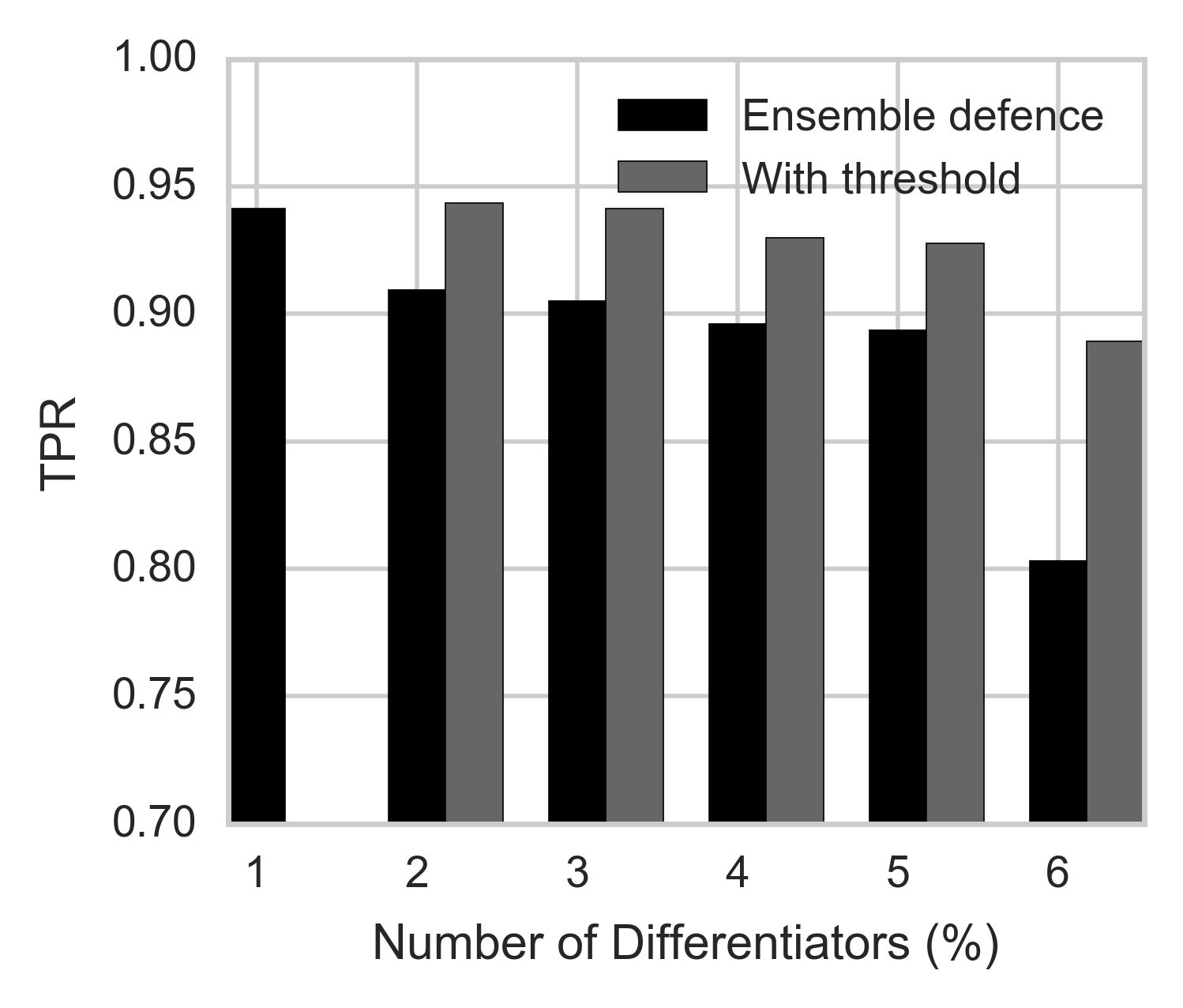}
        
        (a) Face Recognition
        
\end{minipage}
\begin{minipage}[htp]{0.54\linewidth}
        \centering
        \includegraphics[width=0.95\textwidth]{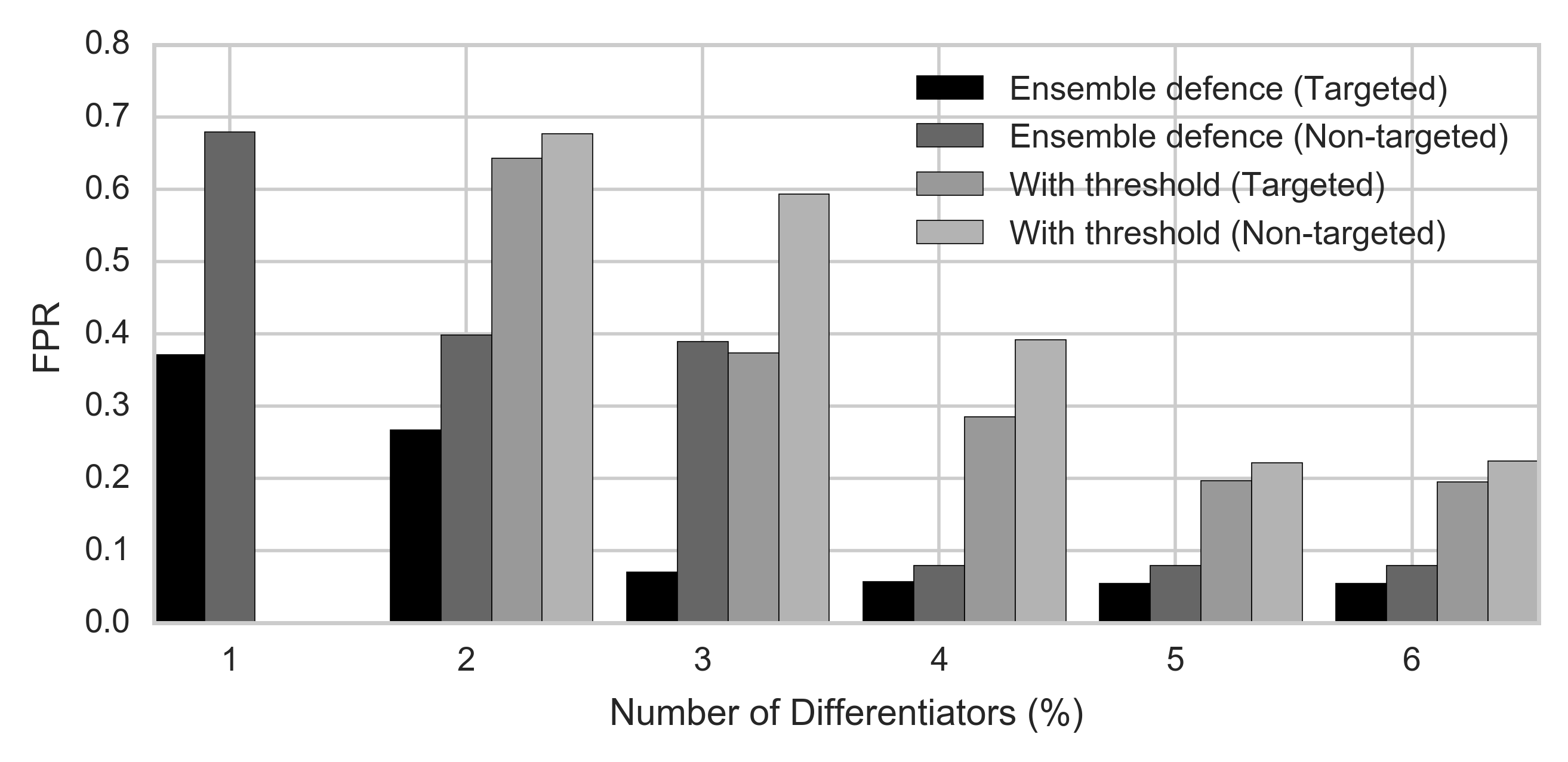}
        
        (b) Traffic Sign Recognition
        
\end{minipage}
\caption{Performance of defence with different numbers of differentiators in Face and Traffic Sign}
\label{fig:Defence_number}
\vspace{-10pt}
\end{figure*}

\subsubsection{Effectiveness}
\noindent 
The effectiveness of our defence is evaluated by both the defence rate and model accuracy.
On the one hand, our models should be able to reject the adversarial inputs. 
So we evaluate our design by the false positive rate (FPR) of the attacker's inputs. 
It equals the number of adversarial examples labelled as benign divides by the number of total adversarial inputs.
Low FPR shows the effectiveness of our defence detecting the adversaries.
On the other hand, the models are expected to classify the benign inputs correctly.
As a result, we also evaluate the accuracy of the models for the clean inputs by using the true positive rate (TPR).
It is defined as the number of inputs validated as positive benign inputs divides by the number of total testing clean inputs.
High TPR means that our defence has less hit on the original classification performance.
Firstly, we compare the performance for the Student models with the defence and the original Student models without any defence.
%
The results confirm that models with our method reject most of the adversarial inputs of the misclassification attacks.

Our defence reduces the FPRs from $100\%$ and $97\%$ for the Face Recognition and Traffic Sign Recognition model to $6\%$ and $1\%$ for the targeted attacks, and $8\%$ and $7\%$ for the non-targeted attacks.
The size of the ensemble models is set to $5$ for both two tasks with $83$ and $43$ classes.
With a single CPU core, the Student model inference costs $0.54$s on average, and the total cost of the same preliminary inference and ensemble latency is around $2.11$s. Due to our small ensemble models, the overhead introduced in the inference does not appear to be the bottleneck in transfer learning systems. 
We also evaluate our defence by applying attacks with different configurations such as a number of differentiators, attack layer, and perturbation budget, to show the scalability of our defence in detail.
It is shown that our design is robust for variants of attacks. 

%

        

        


\noindent \textbf{Number of differentiators:}
We evaluate our defence by gradually increasing the number of differentiators in the second step.
The results show that a small number of differentiators ($5$ in our experiments) used in the second phase of our design can be sufficient to defend against the attacks.

The number of differentiators used in the second phase of our defence will also affect the defence performance. 
Based on our justification in Section~\ref{sec:defence}, a small set of our differentiators are selected for the prediction to balance the trad-offs between the effectiveness (evaluated by FPR) and accuracy (evaluated by TPR) of our defence. 

Figure~\ref{fig:Defence_number} $a$ shows the relationship among the FPR of both targeted and non-targeted adversarial inputs, the TPR and the number of differentiators in the step two inference for the Face Recognition task.
For Deep-layer Feature Extractor, based on the experience of prior work~\cite{Wang18}, the attack success rates to the original Student model without our defence are more than $95\%$ for both the targeted and non-targeted attacks.
After applying our design, about $90\%$ for both targeted and non-targeted attacks are detected and rejected with little accuracy loss after $5$ differentiators inference.
Figure~\ref{fig:Defence_number}$b$ shows the relationship for the Traffic Signs Recognition.
It can be seen that the FPR drops to less than $10\%$ after $4$ differentiators.

We also evaluate how our defence affects the classification of the clean inputs.
Figure~\ref{fig:Defence_number} also shows the TPR. 
As Figure~\ref{fig:Defence_number}$b$ shows, the TPR is evidently reduced after combining more than $5$ differentiators.
It is caused by the false negative when increasing the number of models.
As seen, the small-scale ensemble models can also preserve the accuracy of our design.
Besides, we evaluate setting a threshold when validating or rejecting in the second phase inference.
We introduce this threshold value to reduce the contingency when judging inputs based on only one negative result.
The inputs are rejected only if the number of negative results is greater than the threshold.
Figure~\ref{fig:Defence_number} compares the FPRs and TPRs when setting the threshold as $1$(our baseline design) and increasing it to $2$.
For both tasks, the TPRs are slightly improved for the larger threshold due to the drop of the False Negative.
However, the FPRs that represent the defence effectiveness decrease dramatically.
Therefore, more differentiators should be applied to compensate for these rising FPRs.
To minimise the latency and the number of inference differentiators, our implementation keeps the threshold to be $1$.

\begin{figure*}[!t]
\centering
\begin{minipage}[htp]{0.33\linewidth}
        \centering
        \includegraphics[width=0.95\textwidth, height=0.57\textwidth]{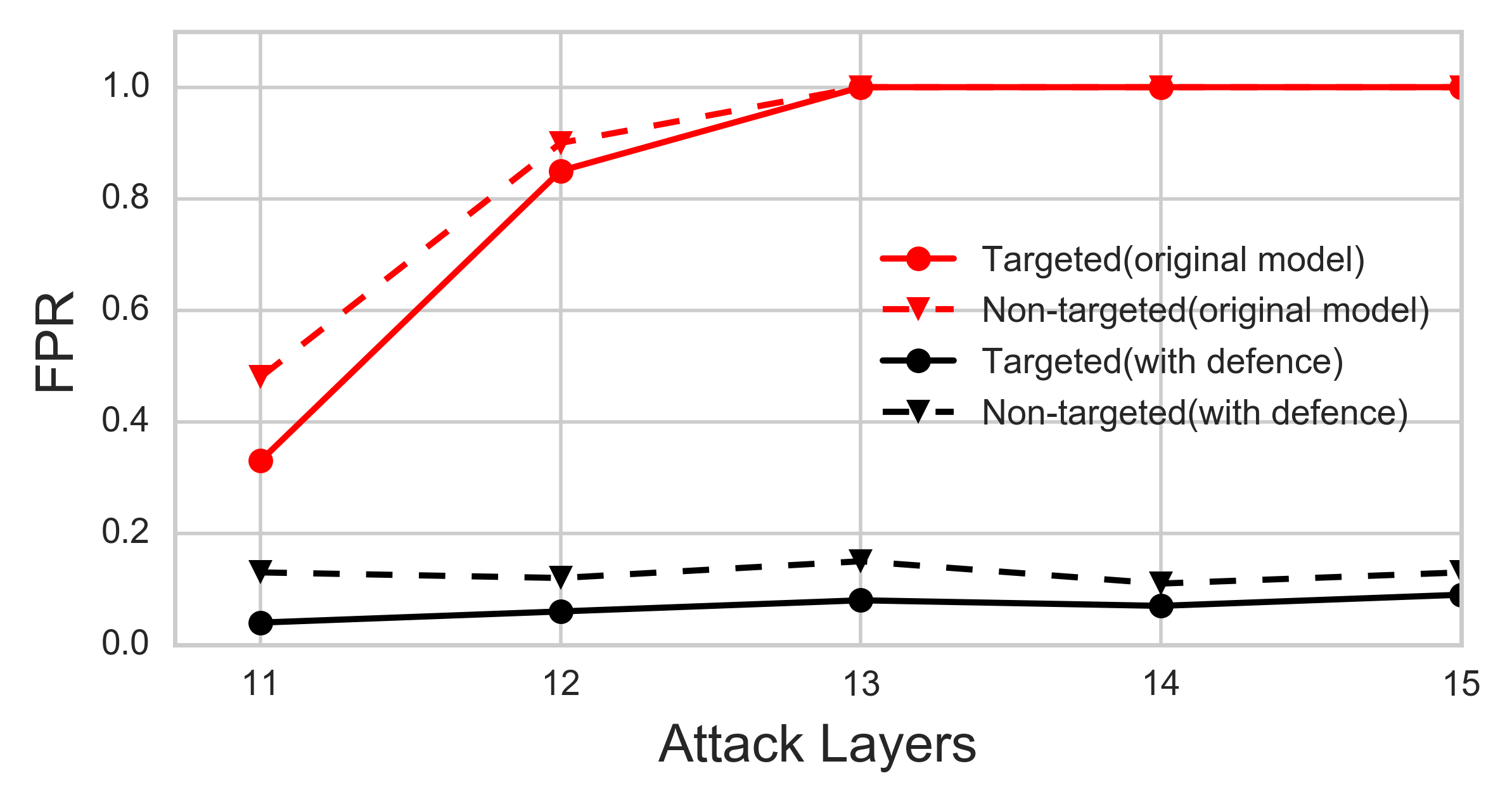}
        
        (a) Face Recognition

\end{minipage}
\begin{minipage}[htp]{0.33\linewidth}
        \centering
        \includegraphics[width=0.95\textwidth, height=0.57\textwidth]{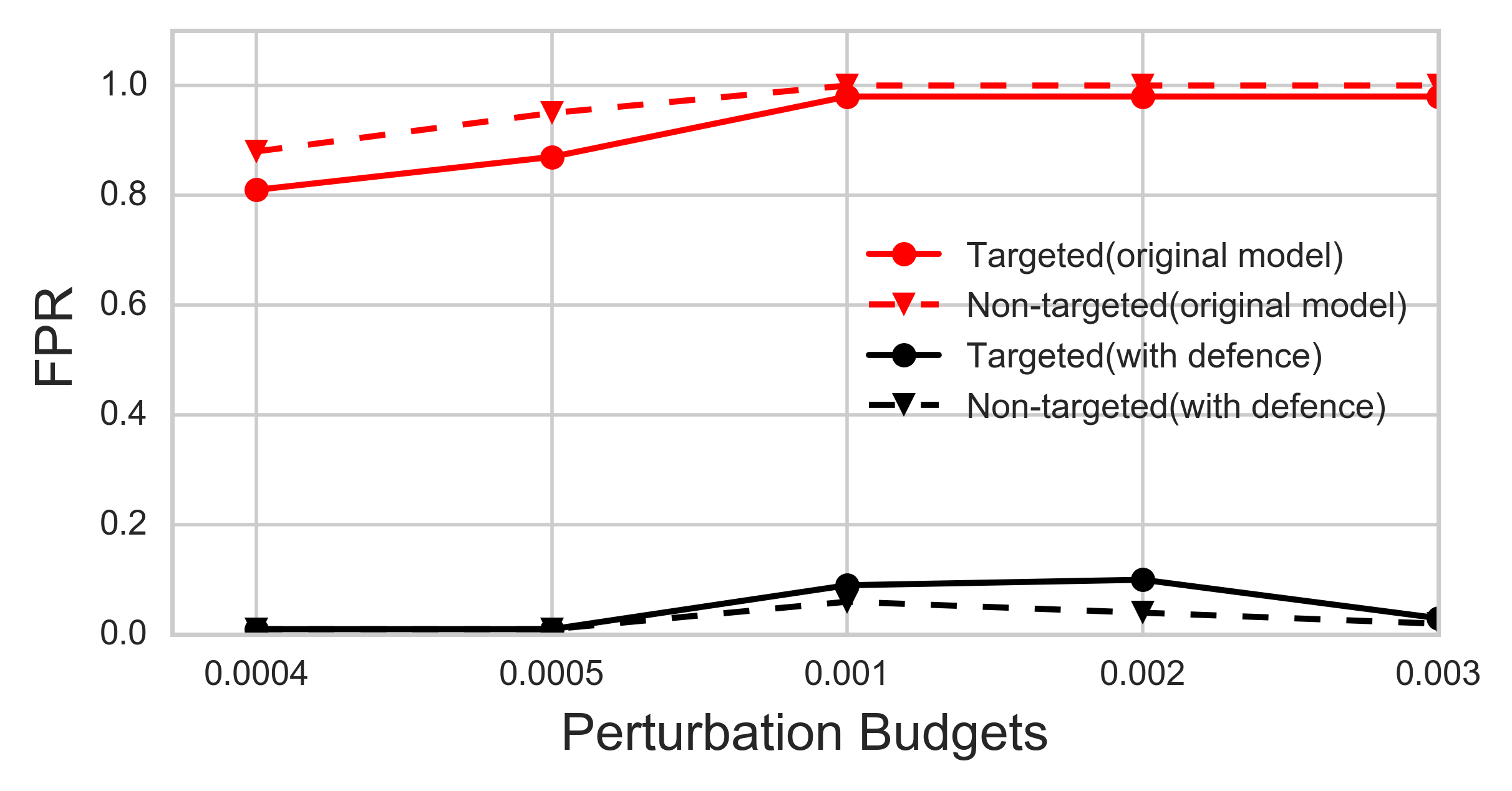}
        
        (a) Face Recognition

\end{minipage}
\begin{minipage}[htp]{0.33\linewidth}
        \centering
        \includegraphics[width=0.95\textwidth, height=0.57\textwidth]{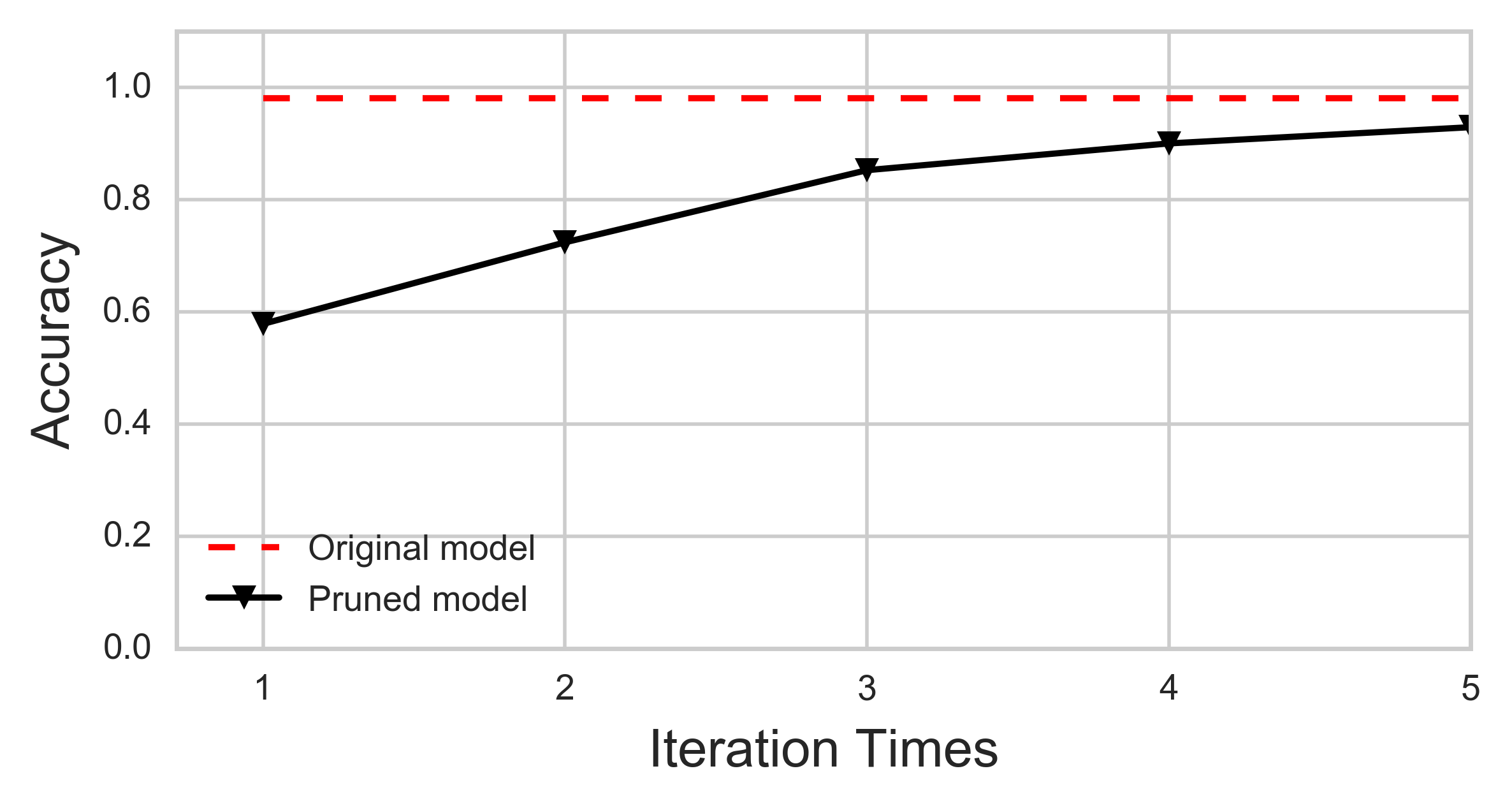}
        
        (a) Face Recognition

\end{minipage}
\\
\begin{minipage}[htp]{0.33\linewidth}
        \centering
        \includegraphics[width=0.95\textwidth, height=0.57\textwidth]{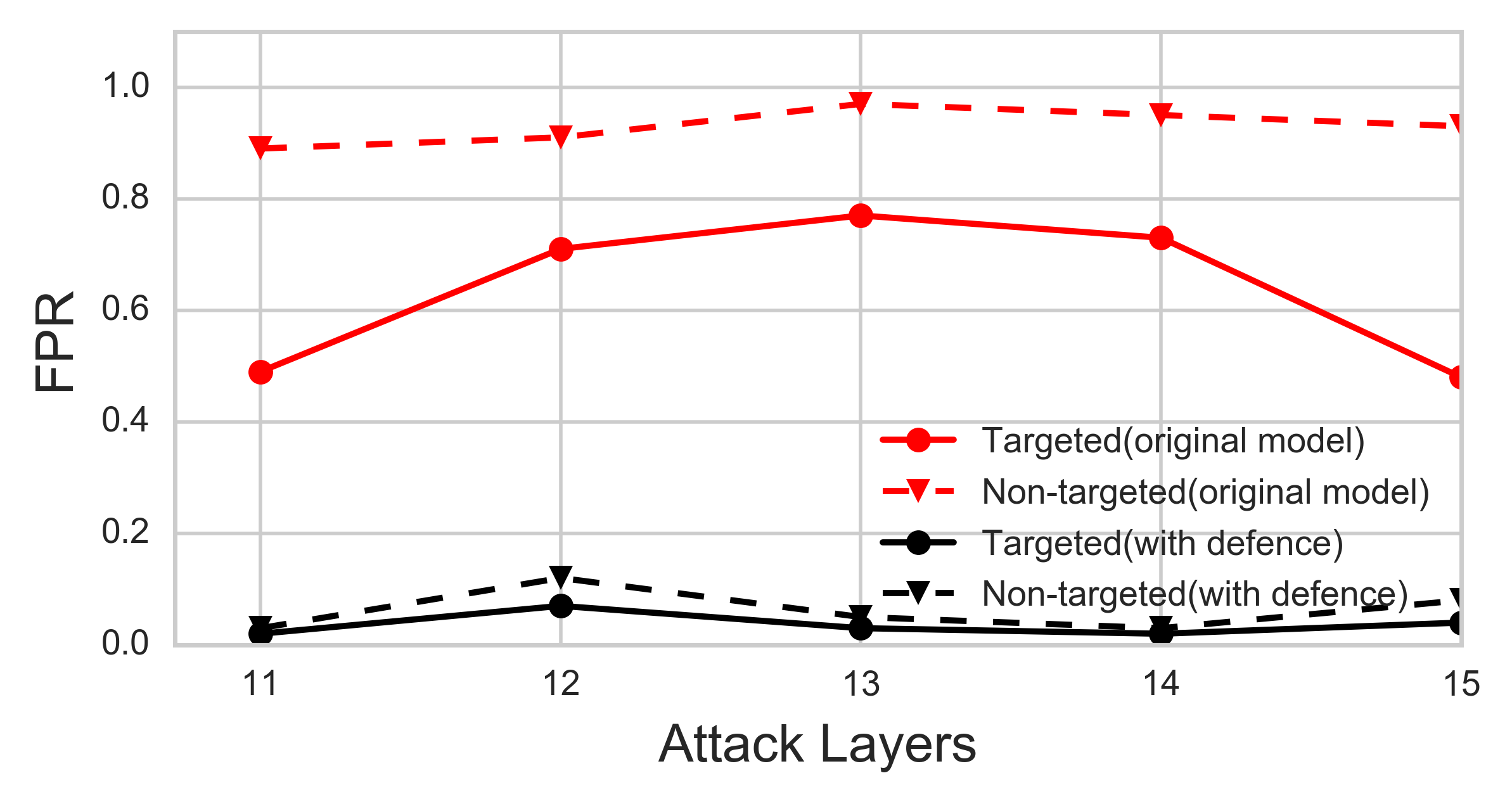}
        
        (b) Traffic Sign Recognition

\caption{Attack Layer}
\label{fig:LayervsAttack}
\end{minipage}
\begin{minipage}[htp]{0.33\linewidth}
        \centering
        \includegraphics[width=0.95\textwidth, height=0.57\textwidth]{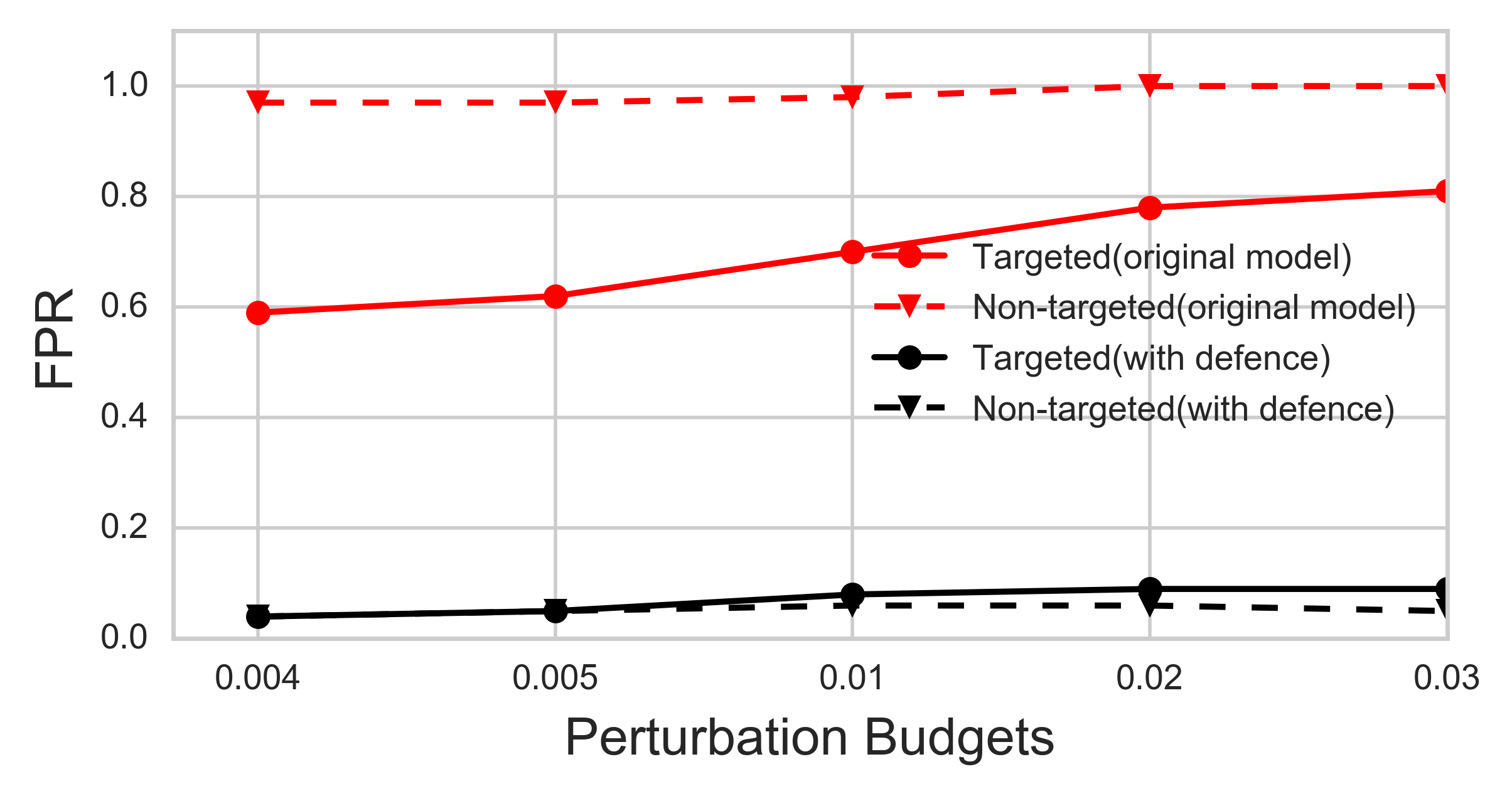}
        
        (b) Traffic Sign Recognition

\caption{Perturbation Budget}
\label{fig:PerturbationvsAttack}
\end{minipage}
\begin{minipage}[htp]{0.33\linewidth}
        \centering
        \includegraphics[width=0.95\textwidth, height=0.57\textwidth]{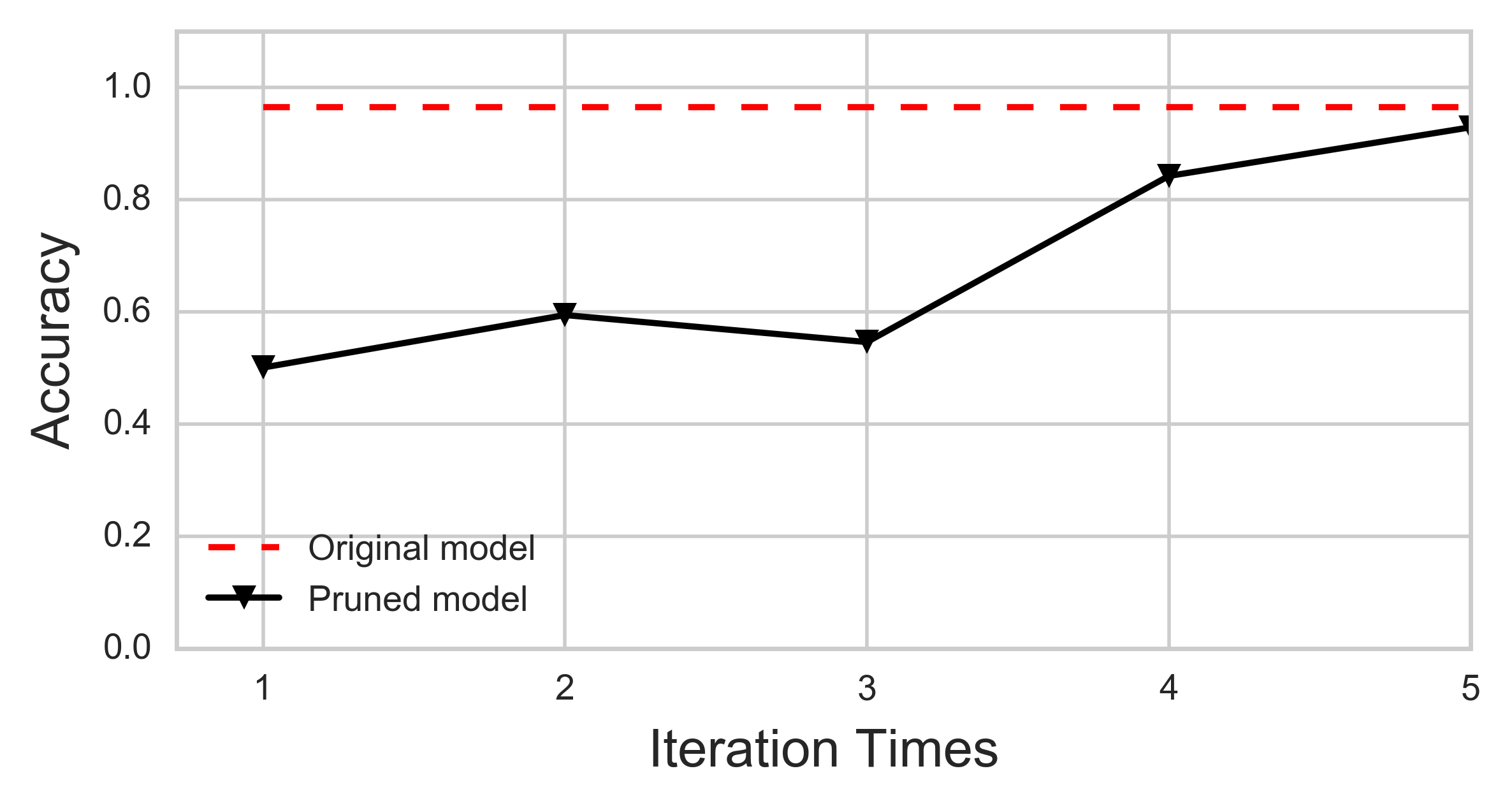}
        
        (b) Traffic Sign Recognition

\caption{Iteration Number}
\label{fig:IterationvsAccuracy}
\end{minipage}
\end{figure*}

\noindent \textbf{Attack Layer:}
We evaluate our defence by applying attacks aiming at different layers.
Since the attackers do not know which transfer method is used for the development of the Student models, they generate small sets of adversarial images targeting several layers to find the optimal attack layer.
Our design is expected to mitigate the attacks aiming at all layers.
%
For Deep-layer Feature Extractor, based on the experience of prior work~\cite{Wang18}, the optimal attack layer is at the last frozen layer.
In Figure~\ref{fig:LayervsAttack}$a$, our defence is quite effective.
For the original Student model without our defence, the success rates for the attack aiming at the last layer are almost $100\%$ for both the targeted and non-targeted attacks.
%
After applying our design, the FPR of the adversarial examples drops to about $4\%$ for the targeted attacks and $9\%$ for the non-targeted attacks. 
Figure~\ref{fig:LayervsAttack}$b$ shows the relationship between the attack layers and the FPR of the adversarial examples with and without the defence for the Traffic Signs Recognition.
%
%
It can be seen that the FPR of the adversarial inputs targeting at different layers keeps small, where most of them are less than $10\%$ for the targeted attacks and $20\%$ for the non-targeted attacks.
Unlike the unguarded models which have an obvious increase of the FPR of the adversarial examples when the attack layer is close to the optimal layer, the variation tendency for our design is flat.
As seen, it is difficult for the attackers to find the optimal layer with a small set of adversarial images. 
In addition, even when the attackers successfully find the optimal attack layer,  the FPR of these adversarial examples is still limited below $20\%$.

        
        
\begin{table*}[!t]
\centering
\begin{tabular}{c|c|c|c|c|c}
\hline
  & Targets & FGSM~\cite{Goodfellow14} & DF~\cite{Moosavi16} & CW~\cite{Carlini17Attack} & EOTPGD~\cite{LiuLWH19}\\ \hline
\multirow{2}{*}{Face}    & Student  & $98.2\%$ & $99.8\%$ & $97.9\%$ & $100.0\%$ \\ \cline{2-6} 
                             & Ensemble       & $6.7\%$ & $2.0\%$ & $3.2\%$ & $31.2\%$ \\ \hline
\multirow{2}{*}{Traffic Sign} & Student & $90.0\%$ & $95.2\%$ & $98.1\%$ & $99.8\%$ \\ \cline{2-6} 
                             & Ensemble & $4.3\%$ & $1.7\%$ & $4.2\%$ & $24.6\%$ \\ \hline
\end{tabular}
\caption{Defending against General Attacks}
\label{tb:DefendGeneralAttacks}
\end{table*}

\begin{table}[h]
\footnotesize
\centering
\begin{tabular}{c|c|c}
 \hline
    &  Unguarded Model & Models w. our design \\
 \hline
 \# of Parameters     &  $1.34 \times 10^8$($512.33MB$) & $1.35 \times 10^8$($513.58MB$) \\
 \hline
    Pruning Tags     & $0$($0MB$)  & $1.20 \times 10^8$($14.25MB$) \\
 \hline
    Total Size & $512.33MB$ & $527.83MB$ \\
 \hline
\end{tabular}
\caption{Memory Consuming of Models with VGG Structure}
\label{tb4:MemoryConsuming1}
\end{table}

\noindent \textbf{Perturbation Budget:}
In practice, the attack configurations like the perturbations added to the inputs can be various.
A larger perturbation budget makes the adversarial inputs stronger to fool the classifiers.
In order to justify our defence's ability in different situations, we evaluate our design for different perturbation budgets. 
%
Figure~\ref{fig:PerturbationvsAttack} shows the relationships between perturbation budgets and the FPR of the adversarial images for both Face and Traffic Sign Recognition.

We choose the optimal attack layers which have the highest attack success rates as shown in Figures~\ref{fig:LayervsAttack}.
According to the results, our design is robust to perturbation.
For the Face Recognition, when the perturbation budgets are in the range of $0.0004$ and $0.003$, the FPR of attacker's inputs targeting unguarded models increases to almost $1$.
And for a larger perturbation budget, the FPR increases observably.
For Traffic Signs Recognition, the attack is effective when the perturbation budget is larger than 0.01.
On the contrary, the models with our defence are more robust to the perturbation variation.
In the test region, the FPRs are less than $10\%$.
\begin{table*}[!t]
\footnotesize
\normalsize
\small
\centering
\begin{tabular}{c|c|c|c|c|c|c}
  \hline
  \multirow{3}{*}{} & \multicolumn{3}{c|}{Face } & \multicolumn{3}{c}{Traffic Sign } \\
\cline{2-7}
~ & \multirow{2}{*}{Accuracy} & \multicolumn{2}{c|}{Attack Success Rate} & \multirow{2}{*}{Accuracy} & \multicolumn{2}{c}{Attack Success  Rate} \\
\cline{3-4}
\cline{6-7}
 ~ & ~ & Targeted & Non-targeted & ~ & Targeted & Non-targeted \\
  \hline
  Original Unguarded Model & $98.1\%$ & $92.5\%$ & $99.8\%$ & $96.5\%$ & $43.5\%$ & $95.2\%$ \\
  \hline
  ETO-based Defence~\cite{PrakashMGDS18} & $96.1\%$ & $90.2\%$ & $91.4\%$ & $94.6\%$ & $33.1\%$ & $91.8\%$ \\
  \hline
  Randomising Input via Dropout~\cite{Wang18} & $49.0\%$ & $4.8\%$ & $90.8\%$ & $90.5\%$ & $2.3\%$ & $40.9\%$ \\
  \hline
  Injecting Neuron Distances~\cite{Wang18} & $95.1\%$ & $20.6\%$ & $79.0\%$ & $94.0\%$ & $8.4\%$ & $62.7\%$ \\
  \hline
  Ensemble Differentiators & $91.6\%$ & $6.1\%$ & $8.5\%$ & $92.9\%$ & $5.6\%$ & $8.1\%$ \\
  \hline
\end{tabular}
\caption{Performance Comparison among Different Defences}
\label{tb3:DefenceComparison1}
\end{table*}


\begin{table*}[!t]
    \centering
    \begin{tabular}{c|c|c|c|c|c|c}
    \hline
     \multicolumn{2}{c|}{ } & \multicolumn{5}{c}{Perturbations} \\ \cline{3-7}
     \multicolumn{2}{c|}{ } & 0.001 & 0.005 & 0.01 & 0.05 & 0.1\\ \hline
    \multirow{4}{*}{Target}    & Unguarded Model  & $98.2\%$ & $98.8\%$ & $98.8\%$ & $100.0\%$ & $100.0\%$  \\ \cline{2-7} 
                                 & Defence Unknown       & $6.1\%$ & $8.2\%$ & $8.9\%$ & $11.8\%$ & $13.2\%$\\ \cline{2-7} 
                                 & Distilled Model Known       & $26.2\%$ & $42.5\%$ & $55.4\%$ & $72.8\%$ & $86.2\%$ \\ \cline{2-7} 
                                 & Pruning Ratios Unknown       & $10.1\%$ & $10.1\%$ & $13.2\%$ & $18.6\%$ & $18.9\%$ \\ \hline
    \multirow{4}{*}{Non-target} & Unguarded Model  & $99.8\%$ & $99.8\%$ & $100.0\%$ & $100.0\%$ & $100.0\%$ \\ \cline{2-7} 
                                 & Defence Unknown       & $8.5\%$ & $9.0\%$ & $17.0\%$ & $32.1\%$ & $32.1\%$ \\ \cline{2-7} 
                                 & Distilled Model Known       & $80.1\%$ & $84.5\%$ & $97.9\%$ & $100.0\%$ & $100.0\%$ \\ \cline{2-7} 
                                 & Pruning Ratios Unknown       & $40.0\%$ & $42.2\%$ & $76.4\%$ & $83.2\%$ & $84.6\%$ \\ \hline
    \end{tabular}
    \caption{Comparison for Different Adversarial Capability.}
    \label{tb6:Difference_Adversarial_Capability1}
    \end{table*}

\noindent \textbf{Extending to Other General Attacks:}
We also evaluate our defence over some common attacks targeting conventional machine learning systems to demonstrate the generality of our design.
We apply FGSM, DeepFool and Carlini-Wagner attacks~\cite{Goodfellow14,Moosavi16,Carlini17Attack} to the Teacher models known to the adversaries and evaluate their adversarial examples by the targeted Student models.
The results show that they are less effective for transfer learning systems where most of their attack success rate drops from almost $100\%$ to about $2\%$.
As we introduced in section~\ref{sec:relatedwork}, these general attacks do lose their effectiveness when the targeted models transfer to new tasks.

We also evaluate the ETO-based attacks in~\cite{LiuLWH19}. 
To extend the attacks to transfer learning scenarios, we assume that the ETO-based attacker constructs their adversarial inputs to cause misclassification of the teacher models. 
The results are presented in Table~\ref{tb:DefendGeneralAttacks}. 
It can be found that, the attack is highly effective when directly attacking the general student models while less effective for our ensemble models. 
Thus, such an attack is less effective in transfer learning scenarios. 
In the adversarial example generation, even though the real-world transformation is taken into account, the attacker is still seeking to increase the loss based on the teacher model rather than the student model. 
As a consequence, when the student models are used for different prediction tasks, the adversarial example generation based on the teacher models will be less effective. 
We also found that, the attack performs better on the Face Recognition task. 
These may be affected by the transfer of learning applications. 
Teacher models and student models for face recognition are both within the same domain, i.e. human faces. 
However, for the traffic sign task, the teacher model focuses on the classification of general objects, while the student model is specific to the classification of traffic signs. 
It becomes more difficult to perform such an attack if the teacher model and the student model have different tasks. 

Note that all these general attacks is less effective in transfer learning scenarios. 
For example, for the ETO-based attacks, in the adversarial example generation, even though the real-world transformation is taken into account, the attacker is still seeking to increase the loss based on the teacher model rather than the student model. 
As a consequence, when the student models are used for different prediction tasks, the adversarial example generation based on the teacher models will be less effective. 

\noindent 
\textbf{Remarks:}
It is important to note that our design is intended for transfer learning, whose settings are quite different from those of many of these attacks. 
These general attacks are often rendered ineffective when the prediction tasks for the student model are different from those for the teacher model.  
In contrast, when an attacker is directly targeting an undefended student model, the study becomes a study of how to defend against evasion attacks in a general case rather than one of transfer learning. 
Such analyses are beyond the scope of this study. 

We further evaluate our design assuming the Student model is known while our defence is unknown to the attackers.
The experiments show that they are also ineffective in this case. The attack success rates of all three attacks drop from above $90\%$ to less than $10\%$ since most of the adversarial examples are rejected by our defence. 
As a result, our design also provides robustness when defending against general attacks. 





\noindent \textbf{Iteration Number:} 
The models applying our design can still maintain the acceptable accuracy for the benign inputs after several iteration periods.
Some defence approaches may affect the performance of classification.
As introduced in Section~\ref{sec:defence}, the original neural networks are pruned in our design. 
Previous studies~\cite{Polyak15,Han15} show that pruning the neural network of classifiers will affect the model accuracy, while iteratively pruning and retraining can help the models regain their accuracy.
In our design, the weights in the frozen layers are fixed, and thus the damage caused by the pruning cannot be recovered by retraining these layers. 
Our experiments show that the iteration pruning and retraining for only the unfrozen layers can still regain an acceptable model accuracy for the clean inputs.
In addition, the pruned models with more iteration numbers for pruning and retraining will lead to higher model accuracy.
%

Figure~\ref{fig:IterationvsAccuracy} illustrates the connection between the iteration times and the accuracy of the Student models for the benign inputs of our two tasks. 
It can be found that the model accuracy for both Face and Traffic Signs Recognition tasks increases to more than $90\%$ after $5$ iterations.
As a result, with iteration pruning and retraining, the accuracy of the ensemble models rises back to an acceptable value. 

\subsubsection{Memory Efficiency Evaluation}


Here, we evaluate the memory consumption of our design.
Table~\ref{tb4:MemoryConsuming1} shows the parameters for the original model and our ensemble models.
In our design, the parameters consist of two parts.
One is the original weights and bias of the Teacher model, which are directly reused in the differentiators.
The memory consumption is the same as a Student model developed by general transfer learning.
Another part is the pruning tags and the individual weights and bias of the last classification layer for each differentiator.
It can be found that most of the parameters of the models are reused.
As seen, our design only consumes extra $3\%$ memory compared to the original.
\subsubsection{Comparison with Others Defences}
%
To further demonstrate our defence's effectiveness, we incorporate a set of defence methods either for mitigating general evasion attacks or specifically design for misclassification attacks. 

\noindent \textbf{Comparison to ETO-based defences:}
We first compare our methodology with defence methods that utilise simple input transformations in~\cite{PrakashMGDS18}
This simple defence has been demonstrated to be effective against several well-known non-adaptive attacks. 
The results have now been summarised in Table~\ref{tb3:DefenceComparison1}. 
We find that this defence method designed for general evasion attacks is less effective when defending against transfer learning attacks especially for the Face Recognition task. 

Since general defence methods are less effective in transfer learning applications, we further compare our defence to the defence specifically for the transfer learning attacks. 
As introduced in Section~\ref{sec:relatedwork}, two basic defence approaches against transfer learning are introduced.
They are \textit{Randomising Input via Dropout} and \textit{Injecting Neuron Distances}.
We further compare our design with these two defences.
The experimental results can be shown in Table~\ref{tb3:DefenceComparison1}.

\noindent \textbf{Comparison to Randomising Input via Dropout:}
In Randomising Input via Dropout, several random pixels of the input images are dropped to decrease the attack success rates of adversarial images.
Although it makes the models more robust, the accuracy of the Student models is severely affected. 
On the contrary, our design maintains much better accuracy of the Student models.
Table~\ref{tb3:DefenceComparison1} reports the comparison between our defence and Randomising Input via Dropout method.
%
For Randomizing Input via Dropout method, the accuracy drops from $98\%$ to $49\%$ after the attacks are being conducted. 
Our design still maintains the classification accuracy at about $92\%$ by doing iteration pruning and retraining.
Moreover, the defence rates of both targeted attacks and non-targeted attacks in our design are higher. 

\noindent \textbf{Comparison to Injecting Neuron Distances:}
Another defence method is Injecting Neuron Distances.
It retrained the whole Student model to increase the distances of the internal feature vectors at the cut-off layer for the inputs.
%
%
The ability of Injecting Neuron Distances defending against the non-targeted attacks is less effective.
In our design, multiple Differentiators are trained for defending against multiple targeted attacks or non-targeted attacks. Such treatment makes our models more robust to these attacks. 
A detailed comparison is shown in Table~\ref{tb3:DefenceComparison1}.
The attack success rate for the non-targeted attacks for Injecting Neuron Distances is about $79\%$ and $63\%$ for two tasks;  it is much higher than our design which is $8.5\%$ and $8.1\%$.


In addition, this method sometimes does not accommodate transfer learning applications with small-scale datasets according to our experiments.
As introduced in Section~\ref{sec:relatedwork}, Injecting Neuron Distances retrains the whole models to increase the dissimilarities of the internal features at the attack layers.
However, this obstacle of training will increase when the targeted Student model has less training data.
In contrast, our defence only fine-tunes a few layers, which still facilitates the strength of transfer learning.


We also compare the time cost of both defence approaches. 
Our experiments with 1 GPU gtx1070 take about 25 minutes to generate a model with injected neuron distance, while about 36 minutes for our design (including all the differentiators rather than only the selected five). 
Training a general student model via transfer learning requires only about 3 minutes since only the unfrozen layers are trained. 
Both the injected neuron distance and our defence significantly increase our development costs. 
However, we would like to emphasise that training the differentiators of our defences requires only retraining the unfrozen layers. 
Hence, even though we train more numbers of models, the training time will be kept at a comparable level to injected neuron distance. 
Moreover, these models can be trained simultaneously so that the development of our defence may be further accelerated, while the other method of tuning the whole student model cannot. 
As the deep learning models become larger, we believe that our design employing similar training strategies as the transfer learning is more suitable. 
Note that this cost is one-time, which is only incurred during setup. 
%



\subsubsection{Adaptive Attack Evaluations}
Based on the guidance principle discussed by Carlini et al. in~\cite{Carlini19}, where the defence algorithm might not be held secret, we enhance the knowledge of the attackers on the defence strategy.

\noindent 
\textbf{Attack Efficiency with Adaptive Attacks:}
Adaptive attacks assume that the attackers are aware of defences and adjust their attack strategies accordingly. 
A common adaptive attack (i.e. attacks introduced in~\cite{TramerCBM20}) dealing with the ensemble models is to modify the objective functions used for the generation of adversarial samples in such a way that the attackers can fool both the general model and all the diverse models. 
Thus, we assume that adaptive attackers, who are aware of our design strategies, will be able to reconstruct all of our differentiators and impose their adaptive attacks. 
We select their attack strategies in ``Theme 5'' of their paper to attack the diverse ensemble, which is more aligned with our defensive policies. 
Note that, to apply such adaptive attacks, we assume a very strong attacker who can reconstruct all the differentiators in a way that allows the attacker to fool the differentiators all by following their proposed strategies. 

\begin{table*}[]
        \centering
        \begin{tabular}{c|cccc|cccc}
        \hline
        \multirow{2}{*}{ } & \multicolumn{8}{c}{Attack Success Rate} \\
        \cline{2-9}
        & \multicolumn{4}{c|}{Face Recognition} & \multicolumn{4}{c}{Traffic Sign Recognition} \\
        \cline{1-9}
           Perturbations & 0.001 & 0.010 & 0.020 & 0.040 & 0.10 & 0.20 & 0.40 & 0.80 \\
        \hline
             Ensemble models & 0.321 & 0.643 & 0.679 & 0.714 & 0.600 & 0.632 & 0.733 & 0.768 \\
             General model & 0.982 & 0.982 & 0.982 & 0.982 & 0.900 & 0.900 & 0.900 & 0.900 \\
        \hline
        \end{tabular}
        \caption{Defending against Adaptive Attacks in~\cite{TramerCBM20}}
        \label{tab:adp_attack_ptb}
    \end{table*}

The evaluations for the attack success rate of the adaptive attack against the general student models and the ensemble models with our defence are shown in Table~\ref{tab:adp_attack_ptb}. 
To find the breaking point and demonstrate the effectiveness of our defence, we increase the perturbations gradually. 
Our results indicate that it is much more difficult for the attacker to attack our ensemble models.
Specifically, a non-adaptive attacker can achieve about $99\%$ attack successful rates to the unguarded student models for Face Recognition, while a more advanced adaptive attacker can only achieve $32\%$ attack successful rates to the models with our defence with the same perturbation budget. 
In addition, more perturbations are needed to generate the adversarial inputs that can bypass our defence comparing with those bypassing the general student models. 
Namely, when the adaptive attackers have been aware of our defence, our defence can still increase the attacker's effort towards infeasibility.  

 Furthermore, we evaluate the memory cost associated with adversarial input generation. 
Namely, we further demonstrate that our defence can also increase the difficulty when an attacker generates adversarial examples. 
The adaptive attacker, as we discussed previously, aims to fool not just one general student model, but all the distilled models in the entire ensemble models. 
These adaptive attacks, for example, using model gradients, will result in increased costs if they are applied to large-scale ensemble models containing many distilled models. 
In Table~\ref{tb:AttackCost} we report the memory cost of the attacker when attacking the original student models and our ensemble models. 

\begin{table*}[t]
    \centering
    \begin{tabular}{c|cc|cc}
     \hline
        & \multicolumn{2}{c|}{Face Recognition} & \multicolumn{2}{c}{Traffic Sign Recognition} \\
     \cline{2-5}
        &  Unguarded Model & Models w. our design &  Unguarded Model & Models w. our design \\
     \hline
     Target Model Size &  $512.33MB$ & $1967.60MB$&  $511.91MB$ & $1769.29MB$ \\
     \hline
     Time Cost & $486.037s$ & $784.604s$ & $423.251s$ & $768.919s$ \\
     \hline
    \end{tabular}
    \caption{Attack Cost per Adversarial Example With/without Our Design. }
    \label{tb:AttackCost}
\end{table*}

\noindent
\textbf{Remark:}
Note that these adaptive attack strategies are not practical in transfer learning scenarios, since they usually require prior knowledge of the targeted models (student models in transfer learning applications). 
Specifically, as an example, the attacks in~\cite{TramerCBM20} modify their objective function when attacking ensemble models so that they can fool all of them at the same time. 
Typically, they must be able to obtain sufficient knowledge from the target models, such as gradients when using white-box settings. 
However, this assumption is not practical in the context of transfer learning. 
In the transfer learning applications, the attackers can have access to only the teacher model but not the student model, including the knowledge of the tasks and dataset. 
As the student models could be very different from the teacher models, it is hard for the attacker to acquire sufficient knowledge based only on the teacher mode then, which makes their attacks become less effective. 

\noindent 
\textbf{Attack Efficiency with Known Defence Methodology:}
To better evaluate our defence under different adversarial capabilities, we assume another attacker who follows the same adversarial examples generation method as the transfer learning attacks (mimic the internal representations), while having different knowledge of the student models. 
Specifically, we consider an adversary who can know our defence strategies, including the activation pruning and the ensemble structure we developed.
Specifically, we assume the attacker can develop exactly the same pruned models as we built.
Namely, the attackers have full access to the Student models and are able to apply the white-box attacks.
The experimental results are also shown in Table~\ref{tb6:Difference_Adversarial_Capability1}. 
With our defence, we can find that the attackers need to have larger perturbation budgets to cause the misclassification. 
Since our differentiators are pruned and highly distilled, it is difficult for the attackers to generate the adversarial examples in limited perturbation budgets.

We also consider a more practical attacker who knows our defence strategies, including the activation pruning and the ensemble structure but not the pruning ratios for each differentiator.
It is reasonable and practical to assume that those ratios are secret parameters and not easily inferred by querying from the attackers due to the ensemble structures of our design.
As a result, it is difficult for the attackers to build the same pruned models exactly as ours, which makes them still hard to attack our design.
The experimental results are also shown in Table~\ref{tb6:Difference_Adversarial_Capability1}. 
With our defence, we can find that the attackers need to have larger perturbation budgets to cause the misclassification. 
Additionally, we found that the attack success rate for targeted attacks remains low (about $10\%$) when attackers are unaware of the defence and do not know the pruning ratios. 
The reason for this is that these attackers lack knowledge of what downstream tasks are in a transfer learning scenario, or that they do not know the exact pruning ratios for each layer. 
Thus, they cannot reconstruct either the student model or the distilled model, which is based on downstream task activation. 
Consequently, it is difficult for them to conduct attacks, especially targeted attacks.

\begin{figure*}[!t]
\centering
\begin{minipage}[htp]{0.45\linewidth}
        \centering
        \includegraphics[width=0.9\textwidth]{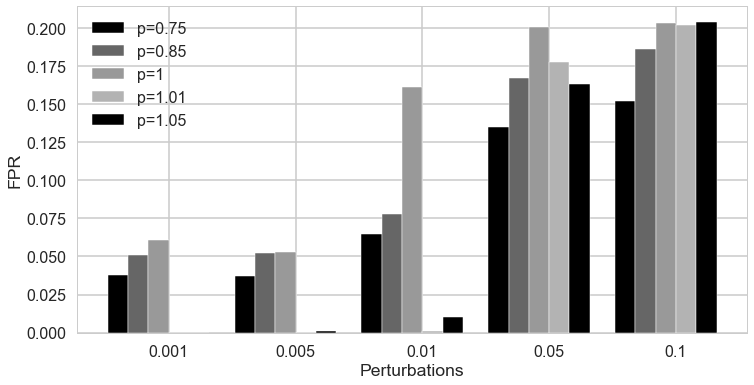}
        
        (a) Targeted Attacks
        
\end{minipage}
\begin{minipage}[htp]{0.45\linewidth}
        \centering
        \includegraphics[width=0.9\textwidth]{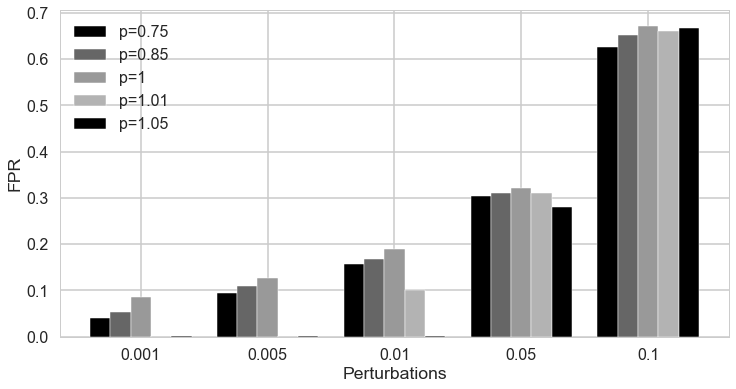}
        
        (a) Non-targeted Attacks
        
\end{minipage}
\caption{Perturbations vs. FPR for Different Pruning Ratios.}
\label{fig:diff_pruning_ratios_ptb}
\end{figure*}

\noindent 
\textbf{Impacts from the Knowledge of Pruning Ratios:}
We further evaluate how well our defence performs when the attacker proposes to generate the adversarial examples with a range of pruning ratios. 
These attackers are assumed to know our pruning strategies as prior knowledge and exploit a value range to cover the exact pruning ratios. 
The results are shown in Figure~~\ref{fig:diff_pruning_ratios_ptb}. 
$p$ is a hyper-parameters that represents the pruning ratio compared to the targeted distilled model. 
For example, $p=0.75$ assume a conservative attacker use the pruning ratios that are only $0.75$ of pruning ratios for the targeted model. 
Based on our observations, the success rate of the non-targeted attacks increases gradually with increases in perturbations, whereas the rate of targeted attacks increases slowly and is limited to under $20\%$. 
We have also observed that, since our models are already fully pruned, when $p$ is greater than $1$, it is very difficult to generate adversarial examples. 
Thus, the attack success rates for the attacker using $p$ exceeding $1$ don't significantly increase until perturbations are greater than $0.05$.
%

\noindent 
\textbf{Remark:}
These the above evaluations are based on the assumption that the attacker already knows the downstream task and they can perform the activation pruning to reproduce a similar distilled model (but by using different pruning ratios). 
However, under transfer learning scenarios, the attackers are commonly assumed not to have such knowledge, in which the effectiveness of the attack will be further diminished when lacking such information. 

\section{Discussion}
\label{sec:discussion}
\begin{table*}[!t]
\footnotesize
\normalsize
\small
\centering
\begin{tabular}{c|c|c|c}
  \hline
  ~ & \multirow{2}{*}{Accuracy} & \multicolumn{2}{c}{Attack Success Rate} \\
  \cline{3-4}
  ~ & ~ & Targeted Attacks & Non-targeted Attacks \\
  \hline
  Original Unguarded Model & $98.1\%$ & $92.5\%$ & $99.8\%$ \\
  \hline
  Differentiator of Two Classes & $98.9\%$ & $9.0\%$ & $13.0\%$ \\
  \hline
  Differentiator of Three Randomly Chosen Classes & $95.0\%$ & $13.5\%$ & $45.0\%$ \\
  \hline
  Differentiator of Three Clustered Classes & $95.0\%$ & $11.8\%$ & $21.5\%$ \\
  \hline
\end{tabular}
\caption{Performance of Multiple-class Differentiator}
\label{tb5:MultiDifferentiator1}
\end{table*}
\noindent \textbf{Classifying Adversarial Images:}
To fit a broader range of applications that require deep analysis on the adversarial inputs, we also provide a design correctly classifying the adversarial images rather than just rejecting them.
Based on our experiments, the differentiators in our design can be highly robust to the adversarial examples whose both source and target class are in their training subset.
Namely, a differentiator trained by a subset consisting of class $i$ and $j$ can correctly classify the adversarial examples that are class $i$ and target at $j$ or the reverse.
Since each differentiator can classify the adversarial examples corresponding to attack pairs, the ensemble models for all of them can be used to classify every possible attack pair.
$K(k-1)/2$ differentiators should be trained in this proposed design. However, this is also a one-time cost during setup.


Considering a classification problem with $K$ classes, there are $\frac{K(K-1)}{2}$ possible label pairs $(c_i,c_j)$, where each of them corresponds to a differentiator.
%
%
%
%
Each of them is trained using their corresponding training subsets.
%
They are expected to be fully pruned and robust to the adversarial inputs according to the design goals of the differentiators.
For any clean inputs, there will always be $K-1$ differentiators which can correctly classify the inputs.
%
Since they are highly robust to the adversarial images, the correct prediction for the inputs can be produced by the voting of the differentiators.


%
The inputs will be identified by all differentiators to get the final predictions where all these predictions will be counted.
Based on our structure, the maximum number of differentiators voting to the same class is $K-1$. 
If $K-1$ differentiators have the same prediction, it is considered as the final outcome.
%
If the maximum number of the differentiators voting to the same class is less than $K-1$, it is considered that some of the differentiators are fooled.
%
As seen, the inputs will be considered as adversarial input and rejected for this case. 

\noindent 
\textbf{Efficiency Improvement:}
The scalability of our design is also considered.
For a complex transfer learning classification task with large numbers of labels, it is expensive to build many such differentiators for each label pair. 
For example, to achieve the best performance, we mentioned that it is better to generate differentiators for each specific attack pair. 
Accordingly, we will need $K(K-1)/2$ differentiators for a $K$ label prediction task. 
It is impractical when the target student model consists of thousands of classes.
So, it is necessary to include multiple classes in a differentiator to improve the efficiency of our design. 
As we mentioned above, building a differentiator for each label pair can make the defence more effective.
If more classes are considered during activation pruning, the difference among the distilled differentiators will decrease, which may affect the successful defence rates.
In order to maintain both differences among the differentiators, we would like the activation of the classes in one differentiator to be similar.
Similar activation will lead to a larger common activated component which can increase the ratios of the unnecessary components of this differentiator.
So it can have larger pruning rate budgets and larger differences to the original and other models.

In order to develop the multiple label differentiator, the clusters are built among all the labels based on their activation.
The labels in one cluster are regarded as similar labels and combined in one differentiator.
And it can reduce the total number of differentiators and make our design more scalable.
General clustering algorithms, such as K-mean, are adequate for our design. 
In practice, we can use other clustering algorithms that relate to activation pruning. 
As an example, an algorithm called ``Agglomerative Class Clustering''~\cite{00070L0020}, in which the labels are grouped according to their activation, may also be applicable to our design. 
In our experiments, we use K-means to cluster the labels. 
We leave other activation-based clustering methods as future work. 

Notice that the number of labels combined in one differentiator is limited.
For a differentiator with more labels, the components activated by these labels also increase, which decreases the pruning rate for accurate models.
Less pruning rate makes them less different from the original Student model, and the transferability of the attacks maintains. 
%
%
%

\noindent \textbf{Multiple Label Pair Differentiator:} 
%
%
In order to reduce the development cost, one initial approach is to combine several differentiators as a multiple label pair differentiator 
%
It targets defending the adversarial images built from more than two classes.
Table~\ref{tb5:MultiDifferentiator1} reports the performance of these differentiators in Face Recognition. As seen, the combination of the differentiators somehow reduces the defence performance in our current experiments.

The attack success rate for the targeted attacks increases to 13\% while for the non-targeted attacks to 45\%.
Since the number of labels increases, more neurons are active during the classification. 
To maintain accuracy, fewer neurons are pruned during the development of our defence. 
This restraint on adversarial images is relieved.
As a result, the defence ability of the design may be affected when multiple differentiators are combined into one. 
Nevertheless, as discussed above, the differentiators consisting of labels chosen by cluster have better performance than randomly chosen labels. 
Table~\ref{tb5:MultiDifferentiator1} compares the performance between both approaches. 
It can be found that, properly choosing the classes can increase the defence rate especially for non-targeted attacks.
\section{Related Work}
\label{sec:relatedwork}
\noindent \textbf{Adversarial Attacks Against Machine Learning Systems:}
%
One of the common adversarial attacks is to build adversarial examples to cause misclassification during inference.
Adversarial examples are (misclassified) inputs to machine learning models but are slightly different from the clean (correctly classified) inputs.
Lots of prior studies show that numbers of machine learning models are vulnerable to these adversarial inputs~\cite{Goodfellow14,Moosavi16,Liu18_asiaccs,Wangtdsc20,Chen20}.
However, it is worth noting that the above general attacks are less effective in transfer learning systems~\cite{Wang18}. 
Recent studies about how to attack the transfer learning system follow two different approaches. 
One tries to demonstrate attacks by generating adversarial models during their development periods~\cite{Ji18}.
Another approach is to generate adversarial inputs for both targeted and non-targeted attacks~\cite{Wang18}.
In practice, it is easier for attackers to use adversarial inputs rather than generating adversarial models during system development~\cite{Goodfellow18}.
Therefore, in this paper, we focus on the latter one proposed by Wang \textit{et al.}~\cite{Wang18} and aim to mitigate the most practical misclassification attacks in current literature.

\noindent \textbf{Defences Against Adversarial Machine Learning:}
Several approaches can increase the robustness of machine learning systems against adversarial examples.
The first is adversarial training, which adds adversarial images to the training datasets during system development~\cite{Kurakin16,Tramer17,ZantedeschiNR17,MadryMSTV18}. 
Others propose to preprocess the inputs before they are sent to the classifiers by extra layers or networks~\cite{MaLTL019,Bendale16,AbbasiG17,Xu0Q18}.
Note that defences such as adversarial training affect the model accuracy, and their implementation is costly, which are less applicable to real-world sensitive applications. 
Recently, Liu \textit{et al.} develop a defence against backdoor attacks via pruning ~\cite{Liu18}.
However, since the structures and weights of the Student Models are changed during development, their defence cannot be directly used for the transfer learning attacks we target~\cite{Wang18}. 
Another research~\cite{WangWYZL18} using the pruning technique implies updating the parameters of the entire pruned model. 
Note that, updating the entire model will significantly increase the difficulty of training and counteract the benefits of transfer learning. 
Therefore, they are not suitable for transfer learning scenarios.

We have not found effective yet efficient methods of defeating the transfer learning misclassification attacks~\cite{Wang18}.
Wang \textit{et al.} suggest two basic approaches which either suffer from accuracy loss or expensive cost.
One is called Randomising Input via Dropout which applies dropout for the input layer of the Models while affecting the model accuracy.
%
Another is Injecting Neuron Distances, which proposes to break the effectiveness of the attacks by increasing the dissimilarities of the internal features at attack layers.
The parameters of the whole model are retrained, which requires large computation costs, especially for large and deep neural networks, and reduces the advantages of transfer learning.
Besides, the non-targeted attacks can still be effective to the above approaches, as demonstrated in~\cite{Wang18}.

As a general defence philosophy, ensemble models have been exploited in~\cite{Bagnall17,Kariyappa19,Liu19,PangXDCZ19}.
However, they can hardly address the attacks in transfer learning scenarios.
Bagnall's defence~\cite{Bagnall17} losses accuracy in the sophisticated models which are the common targets for transfer learning.
Others~\cite{Kariyappa19,Liu19,PangXDCZ19} retrain and update the entire network for each ensemble model based on different strategies ,such as adaptive diversity promoting training (ADP). 
However, they are less effective in transfer learning scenarios where the developer only has limited data and computation resources. 
Our design employs customised approaches in the transfer learning system, which overcomes the limitations of the above methods.
A recent study~\cite{He17} observes that assembling several weak defences is not sufficient to develop a strong defence. We note that their results are different from our scenarios and not applicable to our design. 
In our design, it is the distilled differentiator which reduces the attack transferability.
Each of our differentiators is considerably effective against the corresponding attacks, and this defence for the specific attacks between two classes is extended to all possible attacks by using the ensemble structure. 
In other words, we effectuate the strong defence of the differentiators rather than combining weak defence by ensemble methods.

\section{Conclusion}
\label{sec:conclusion}
In this paper, we propose and implement a practical defence against the strong misclassification attacks in transfer learning applications.
We use network pruning to develop distilled differentiators that reduce the transferability of the attacks and improve their robustness against adversarial examples.
Activation pruning and flexible pruning ratio selection are applied to preserve accuracy.
In addition, we introduce ensemble methods to further improve the robustness of our design based on a two-phase inference constructed by the Student model and a small-scale of two-class differentiators.
We demonstrate that the ensemble models can effectively defend against both the targeted and non-targeted attacks by rejecting most of the adversarial examples. 
We also show that our design preserves the scalability, effectiveness and comparable clean input accuracy with a small size of ensemble models.
Besides, we further evaluate our defence compared to other defence approaches.
Our design is shown to be more effective and accessible to be implemented.
\section*{Acknowledgment}
This work was supported in part by a Monash-Data61 collaborative research project under Data61 CRP43, and the Research Grants Council of Hong Kong under Grant CityU 11217819, N\_CityU139/21, and R6021-20F


\ifCLASSOPTIONcaptionsoff
  \newpage
\fi



%


\bibliographystyle{IEEEtran}
\bibliography{aml}

\begin{thebibliography}{10}
\providecommand{\url}[1]{#1}
\csname url@samestyle\endcsname
\providecommand{\newblock}{\relax}
\providecommand{\bibinfo}[2]{#2}
\providecommand{\BIBentrySTDinterwordspacing}{\spaceskip=0pt\relax}
\providecommand{\BIBentryALTinterwordstretchfactor}{4}
\providecommand{\BIBentryALTinterwordspacing}{\spaceskip=\fontdimen2\font plus
\BIBentryALTinterwordstretchfactor\fontdimen3\font minus
  \fontdimen4\font\relax}
\providecommand{\BIBforeignlanguage}[2]{{%
\expandafter\ifx\csname l@#1\endcsname\relax
\typeout{** WARNING: IEEEtran.bst: No hyphenation pattern has been}%
\typeout{** loaded for the language `#1'. Using the pattern for}%
\typeout{** the default language instead.}%
\else
\language=\csname l@#1\endcsname
\fi
#2}}
\providecommand{\BIBdecl}{\relax}
\BIBdecl

\bibitem{GoogleAI}
{Google}, ``{Google Cloud AutoML},'' Online at
  \url{https://cloud.google.com/automl/}, 2019.

\bibitem{ChenZZ0S020}
Y.~Chen, B.~Zheng, Z.~Zhang, Q.~Wang, C.~Shen, and Q.~Zhang, ``Deep learning on
  mobile and embedded devices: State-of-the-art, challenges, and future
  directions,'' \emph{{ACM} Comput. Surv.}, 2020.

\bibitem{DengJ09}
J.~Deng, W.~Dong, R.~Socher, L.~Li, K.~Li, and F.~Li, ``Imagenet: {A}
  large-scale hierarchical image database,'' in \emph{Proc. {CVPR} 2009}.

\bibitem{Szegedy16}
C.~Szegedy, V.~Vanhoucke, S.~Ioffe, J.~Shlens, and Z.~Wojna, ``Rethinking the
  inception architecture for computer vision,'' in \emph{Proc. {CVPR} 2016}.

\bibitem{Wang18}
B.~Wang, Y.~Yao, B.~Viswanath, H.~Zheng, and B.~Y. Zhao, ``With great training
  comes great vulnerability: Practical attacks against transfer learning,'' in
  \emph{Proc. {USENIX} Security 2018}.

\bibitem{Ji18}
Y.~Ji, X.~Zhang, S.~Ji, X.~Luo, and T.~Wang, ``Model-reuse attacks on deep
  learning systems,'' in \emph{Proc. {CCS} 2018}.

\bibitem{Xu0Q18}
W.~Xu, D.~Evans, and Y.~Qi, ``Feature squeezing: Detecting adversarial examples
  in deep neural networks,'' in \emph{Proc. {NDSS}, 2018}.

\bibitem{MaLTL019}
S.~Ma, Y.~Liu, G.~Tao, W.~Lee, and X.~Zhang, ``{NIC:} detecting adversarial
  samples with neural network invariant checking,'' in \emph{Proc. {NDSS},
  2019}.

\bibitem{Tramer17}
F.~Tram{\`{e}}r, A.~Kurakin, N.~Papernot, I.~J. Goodfellow, D.~Boneh, and P.~D.
  McDaniel, ``Ensemble adversarial training: Attacks and defenses,'' in
  \emph{Proc. {ICLR} 2018}.

\bibitem{Kurakin16}
A.~Kurakin, I.~J. Goodfellow, and S.~Bengio, ``Adversarial machine learning at
  scale,'' in \emph{Proc. {ICLR} 2017}.

\bibitem{Goodfellow14}
I.~J. Goodfellow, J.~Shlens, and C.~Szegedy, ``Explaining and harnessing
  adversarial examples,'' in \emph{Proc. {ICLR} 2015}.

\bibitem{Moosavi16}
S.~Moosavi{-}Dezfooli, A.~Fawzi, and P.~Frossard, ``Deepfool: {A} simple and
  accurate method to fool deep neural networks,'' in \emph{Proc. {CVPR} 2016}.

\bibitem{Carlini19}
N.~Carlini, A.~Athalye, N.~Papernot, W.~Brendel, J.~Rauber, D.~Tsipras,
  I.~Goodfellow, A.~Madry, and A.~Kurakin, ``On evaluating adversarial
  robustness,'' \emph{arXiv preprint arXiv:1902.06705}, 2019.

\bibitem{Han15}
S.~Han, J.~Pool, J.~Tran, and W.~J. Dally, ``Learning both weights and
  connections for efficient neural network,'' in \emph{Proc. NIPS, 2015}.

\bibitem{Wang19}
B.~Wang, Y.~Yao, S.~Shan, H.~Li, B.~Viswanath, H.~Zheng, and B.~Y. Zhao,
  ``Neural cleanse: Identifying and mitigating backdoor attacks in neural
  networks,'' in \emph{Proc. {IEEE SP} 2019}.

\bibitem{LiuSZHD19}
Z.~Liu, M.~Sun, T.~Zhou, G.~Huang, and T.~Darrell, ``Rethinking the value of
  network pruning,'' in \emph{Proc. {ICLR} 2019}.

\bibitem{MolchanovAV17}
D.~Molchanov, A.~Ashukha, and D.~P. Vetrov, ``Variational dropout sparsifies
  deep neural networks,'' in \emph{Proc. {ICML} 2017}.

\bibitem{Polyak15}
A.~Polyak and L.~Wolf, ``Channel-level acceleration of deep face
  representations,'' \emph{{IEEE} Access}, 2015.

\bibitem{Liu18}
K.~Liu, B.~Dolan{-}Gavitt, and S.~Garg, ``Fine-pruning: Defending against
  backdooring attacks on deep neural networks,'' in \emph{Proc. {RAID} 2018}.

\bibitem{Li16}
H.~Li, A.~Kadav, I.~Durdanovic, H.~Samet, and H.~P. Graf, ``Pruning filters for
  efficient convnets,'' in \emph{Proc. {ICLR} 2017}.

\bibitem{Pinto11}
N.~Pinto, Z.~Stone, T.~E. Zickler, and D.~D. Cox, ``Scaling up
  biologically-inspired computer vision: {A} case study in unconstrained face
  recognition on facebook,'' in \emph{Proc. {CVPR} Workshops 2011}.

\bibitem{Stallkamp11}
J.~Stallkamp, M.~Schlipsing, J.~Salmen, and C.~Igel, ``The german traffic sign
  recognition benchmark: {A} multi-class classification competition,'' in
  \emph{Proc. {IJCNN} 2011}.

\bibitem{Parkhi15}
O.~M. Parkhi, A.~Vedaldi, and A.~Zisserman, ``Deep face recognition,'' in
  \emph{Proc. {BMVC} 2015}.

\bibitem{kumar09}
N.~Kumar, A.~C. Berg, P.~N. Belhumeur, and S.~K. Nayar, ``Attribute and simile
  classifiers for face verification,'' in \emph{Proc. {ICCV} 2009}.

\bibitem{Simonyan14}
K.~Simonyan and A.~Zisserman, ``Very deep convolutional networks for
  large-scale image recognition,'' in \emph{Proc. {ICLR} 2015}.

\bibitem{Stallkamp12}
J.~Stallkamp, M.~Schlipsing, J.~Salmen, and C.~Igel, ``Man vs. computer:
  Benchmarking machine learning algorithms for traffic sign recognition,''
  \emph{Neural Networks}, 2012.

\bibitem{Zeiler12}
M.~D. Zeiler, ``{ADADELTA:} an adaptive learning rate method,'' \emph{CoRR},
  vol. abs/1212.5701, 2012.

\bibitem{Carlini17Attack}
N.~Carlini and D.~A. Wagner, ``Towards evaluating the robustness of neural
  networks,'' in \emph{Proc. {IEEE SP} 2017}.

\bibitem{LiuLWH19}
X.~Liu, Y.~Li, C.~Wu, and C.~Hsieh, ``Adv-bnn: Improved adversarial defense
  through robust bayesian neural network,'' in \emph{Proc. {ICLR} 2019}.

\bibitem{PrakashMGDS18}
A.~Prakash, N.~Moran, S.~Garber, A.~DiLillo, and J.~A. Storer, ``Deflecting
  adversarial attacks with pixel deflection,'' in \emph{Proc. {CVPR} 2018}.

\bibitem{TramerCBM20}
F.~Tram{\`{e}}r, N.~Carlini, W.~Brendel, and A.~Madry, ``On adaptive attacks to
  adversarial example defenses,'' in \emph{Proc. NeurIPS 2020}.

\bibitem{00070L0020}
Y.~Li, M.~Li, B.~Luo, Y.~Tian, and Q.~Xu, ``Deepdyve: Dynamic verification for
  deep neural networks,'' in \emph{Proc. {CCS} 2020}.

\bibitem{Liu18_asiaccs}
S.~Liu, J.~Zhang, Y.~Wang, W.~Zhou, Y.~Xiang, and O.~D. Vel., ``A data-driven
  attack against support vectors of svm,'' in \emph{Proc ASIACCS 2018}.

\bibitem{Wangtdsc20}
D.~Wang, C.~Li, S.~Wen, S.~Nepal, and Y.~Xiang, ``Man-in-the-middle attacks
  against machine learning classifiers via malicious generative models,''
  \emph{IEEE Transactions on Dependable and Secure Computing}, 2020.

\bibitem{Chen20}
Y.~{Chen}, C.~{Shen}, C.~{Wang}, Q.~{Xiao}, K.~{Li}, and Y.~{Chen}, ``Scaling
  camouflage: Content disguising attack against computer vision applications,''
  \emph{IEEE Transactions on Dependable and Secure Computing}, 2020.

\bibitem{Goodfellow18}
I.~J. Goodfellow, P.~D. McDaniel, and N.~Papernot, ``Making machine learning
  robust against adversarial inputs,'' \emph{Commun. {ACM}}, 2018.

\bibitem{ZantedeschiNR17}
V.~Zantedeschi, M.~Nicolae, and A.~Rawat, ``Efficient defenses against
  adversarial attacks,'' in \emph{Proc. AISec@CCS 2017}.

\bibitem{MadryMSTV18}
A.~Madry, A.~Makelov, L.~Schmidt, D.~Tsipras, and A.~Vladu, ``Towards deep
  learning models resistant to adversarial attacks,'' in \emph{Proc. {ICLR}
  2018}.

\bibitem{Bendale16}
A.~Bendale and T.~E. Boult, ``Towards open set deep networks,'' in \emph{Proc.
  {CVPR} 2016}.

\bibitem{AbbasiG17}
M.~Abbasi and C.~Gagn{\'{e}}, ``Robustness to adversarial examples through an
  ensemble of specialists,'' in \emph{Proc. {ICLR} 2017}.

\bibitem{WangWYZL18}
S.~Wang, X.~Wang, S.~Ye, P.~Zhao, and X.~Lin, ``Defending {DNN} adversarial
  attacks with pruning and logits augmentation,'' in \emph{Proc. GlobalSIP
  2018}.

\bibitem{Bagnall17}
A.~Bagnall, R.~C. Bunescu, and G.~Stewart, ``Training ensembles to detect
  adversarial examples,'' \emph{CoRR}, 2017.

\bibitem{Kariyappa19}
S.~Kariyappa and M.~K. Qureshi, ``Improving adversarial robustness of ensembles
  with diversity training,'' \emph{CoRR}, 2019.

\bibitem{Liu19}
L.~Liu, W.~Wei, K.~H. Chow, M.~Loper, M.~E. Gursoy, S.~Truex, and Y.~Wu, ``Deep
  neural network ensembles against deception: Ensemble diversity, accuracy and
  robustness,'' in \emph{Proc. {MASS} 2019}.

\bibitem{PangXDCZ19}
T.~Pang, K.~Xu, C.~Du, N.~Chen, and J.~Zhu, ``Improving adversarial robustness
  via promoting ensemble diversity,'' in \emph{Proc. {ICML} 2019}.

\bibitem{He17}
W.~He, J.~Wei, X.~Chen, N.~Carlini, and D.~Song, ``Adversarial example defense:
  Ensembles of weak defenses are not strong,'' in \emph{Proc. {USENIX} Workshop
  on Offensive Technologies, 2017}.

\end{thebibliography}

\end{document}